\documentclass{bmvc2k}

\title{Evaluating Video LLMs' Understanding of\\ Corner Cases in Autonomous Driving}

\addauthor{Gabriel Manalu}{gabrieldavin23@kaist.ac.kr} \\
\addauthor{Minh Son Hoang}{sonhm2910@kaist.ac.kr} \\
\addauthor{Hyeontaek Hwang}{hwanght@kaist.ac.kr} \\
\addauthor{Jaesung Lee}{ljs6017@kaist.ac.kr} \\
\addauthor{Hyojin Choi}{hyojinchoi@kaist.ac.kr} \\
\addauthor{Tran Tuan Phong}{phongtt263@kaist.ac.kr} \\
\addauthor{Seon Kwon Kim}{lukaskim@kaist.ac.kr} \\
\addauthor{Saad Wazir}{saad.wazir@kaist.ac.kr} \\
\addauthor{Daeyoung Kim}{kimd@kaist.ac.kr} \\

\addinstitution{
 Korea Advanced Institute of Science and Technology \\
 Daejeon, Korea
}

\runninghead{Manalu et al}{Evaluating Video LLM corner case understanding}

\usepackage{multirow}
\usepackage{enumitem}
\usepackage{graphicx}
\usepackage{xcolor}
\usepackage{soul} %
\usepackage{tikz}
\usepackage{caption}
\global
\usepackage{pifont}
\usepackage{xspace}
\usepackage{tabularx} %
\usepackage{rotating} %
\usepackage{wrapfig}
\usepackage{pifont}
\usepackage{colortbl}
\usepackage{float}
\usepackage{comment}
\usepackage[skins,breakable]{tcolorbox}

\newcommand{\oThree}{\texttt{o3}\xspace}

\newcommand{\deepseekvthree}{\texttt{DeepSeekV3}\xspace}
\newcommand{\llama}{\texttt{Llama3.1(70B)}\xspace}

\newcommand{\gptfive}{\texttt{GPT-5}\xspace}
\newcommand{\geminiflash}{\texttt{Gemini-3-Flash}\xspace}

\newcommand{\llavaonevision}{\texttt{LLaVA-OV-1.5}\xspace}
\newcommand{\qwenvl}{\texttt{Qwen3-VL}\xspace}
\newcommand{\ovistwofive}{\texttt{Ovis2.5}\xspace}
\newcommand{\keyevl}{\texttt{KeyeVL}\xspace}
\newcommand{\internvlthreefive}{\texttt{InternVL3.5}\xspace}
\newcommand{\internvideo}{\texttt{InternVideo2.5}\xspace}
\newcommand{\minicpmfourfive}{\texttt{MiniCPM4.5}\xspace}
\newcommand{\glmfouronevthinking}{\texttt{GLM-4.1V-Thinking}\xspace}
\newcommand{\llavanextvideo}{\texttt{LLaVA-Next-Video}\xspace}
\newcommand{\tarsiertwo}{\texttt{Tarsier2}\xspace}
\newcommand{\vitaonefive}{\texttt{VITA-1.5}\xspace}
\newcommand{\internlmxcomposertwo}{\texttt{IXC-2.5}\xspace}

\newcommand{\drivemm}{\texttt{DriveMM}\xspace}
\newcommand{\dolphins}{\texttt{Dolphins}\xspace}
\newcommand{\miniinternvl}{\texttt{MiniInternVL2}\xspace}
\newcommand{\miniinternvldrivelm}{\texttt{MiniInternVL2-DLM}\xspace}
\newcommand{\miniinternvlbdd}{\texttt{MiniInternVL2-BDD}\xspace}
\newcommand{\llavaroadsocial}{\texttt{LlaVA-OV-RS}\xspace}
\newcommand{\drivelmm}{\texttt{DriveLMM-o1}\xspace}
\newcommand{\crashchat}{\texttt{CrashChat}\xspace}

\newcommand{\cmark}{\textcolor{green!70!black}{\ding{51}}}
\newcommand{\xmark}{\textcolor{red}{\ding{55}}}

\newcommand{\pplus}[1]{{\scriptstyle\textcolor{green!55!black}{+#1}}}
\newcommand{\pminus}[1]{{\scriptstyle\textcolor{red!80!black}{-#1}}}

\newcommand{\qone}{Q1 (\texttt{What is unusual in this sequence of frames?})\xspace}
\newcommand{\qtwo}{Q2 (\texttt{What is dangerous in this sequence of frames?})\xspace}
\newcommand{\qthree}{Q3 (\texttt{Describe this sequence of frames in detail.})\xspace}

\newcommand{\qfive}{\texttt{A corner case is an unexpected or unusual driving situation that deviates from normal traffic  behavior, often rare and potentially hazardous. Corner cases may appear as visual anomalies, environmental shifts, unexpected objects, unusual object arrangements, or rare dynamic events.  Examples include a pedestrian suddenly running onto the road from behind an occlusion, a car running a red light, or dropped cargo on the street.\newline\newline Here is a sequence of frames from a dashcam video. Identify if there is a corner case in this dashcam  video, and describe the corner case you identified.}\xspace}

\tcbset{
  aibox/.style={
    width=\textwidth,         
    top=8pt,                      
    colback=white,
    colframe=black,
    colbacktitle=black,
    boxsep      = 2pt,               %
    top         = 2pt,               %
    bottom      = 2pt,               %
    titlerule  = 0mm,               %
    fonttitle=\small, 
    fontupper=\small,
    enhanced,
    before skip=8pt, after skip=8pt,
  }
}
\newtcolorbox{AIboxFull}[2][]{aibox, title=#2, #1}

\tcbset{
  aiboxhalf/.style={
    width=\linewidth,
    colback=white,
    colframe=black,
    colbacktitle=black,
    fonttitle=\scriptsize,
    fontupper=\scriptsize,
    boxsep=2pt,
    left=1pt,
    right=1pt,
    top=0.5pt,
    bottom=2pt,
    titlerule=0mm,
    before skip=8pt,
    after skip=8pt,
  }
}
\newtcolorbox{AIboxHalf}[2][]{aiboxhalf, title=#2, #1}

\tcbset{
  aiboxmain/.style={
    width=\textwidth,                              
    colback=white,
    colframe=black,
    colbacktitle=black,
    fonttitle=\scriptsize, 
    fontupper=\tiny,           
    enhanced,
    boxsep      = 2pt,               %
    top         = 2pt,               %
    bottom      = 2pt,               %
    titlerule  = 0mm,               %
    before skip=8pt, after skip=8pt,
  }
}
\newtcolorbox{AIboxMain}[2][]{aiboxmain, title=#2, #1}

\begin{document}

\maketitle

\begin{abstract}
Natural language explanations promised by modern Video Large Language Models (Video LLMs) offer autonomous driving systems a path toward detailed scene description and interpretable decision making on the road. While recent benchmarks have begun assessing these models on driving scenes, they mainly target routine traffic conditions and general comprehension tasks based on existing driving datasets and therefore fail to evaluate model performance on the rarest and most safety‑critical driving events, also known as "\emph{corner cases}". To provide a comprehensive evaluation of Video LLMs on precisely these scenarios, we introduce \textbf{DashBench}, a benchmark testing 25 Video LLMs on 1,709 carefully curated driving videos spanning not only collision-related corner cases but also rare non-collision ones. Through extensive prompts and open‑ended question answering, we expose a critical capability gap: all evaluated models perform substantially below the threshold required for safety‑critical driving applications, frequently misjudging or missing dangerous events on the road ahead and hallucinating non‑existent agents or actions in their explanations. We believe \textbf{DashBench}, combining an extensive dataset with a rigorous evaluation toolkit and a detailed diagnosis of current model limitations, drives progress toward safer, more explainable autonomous driving systems. The code and data will be shared at: \href{https://cornercase-dashbench.github.io}{cornercase-dashbench.github.io}
\end{abstract}

\begin{figure}[H]
    \centering
    \includegraphics[width=0.87\linewidth]{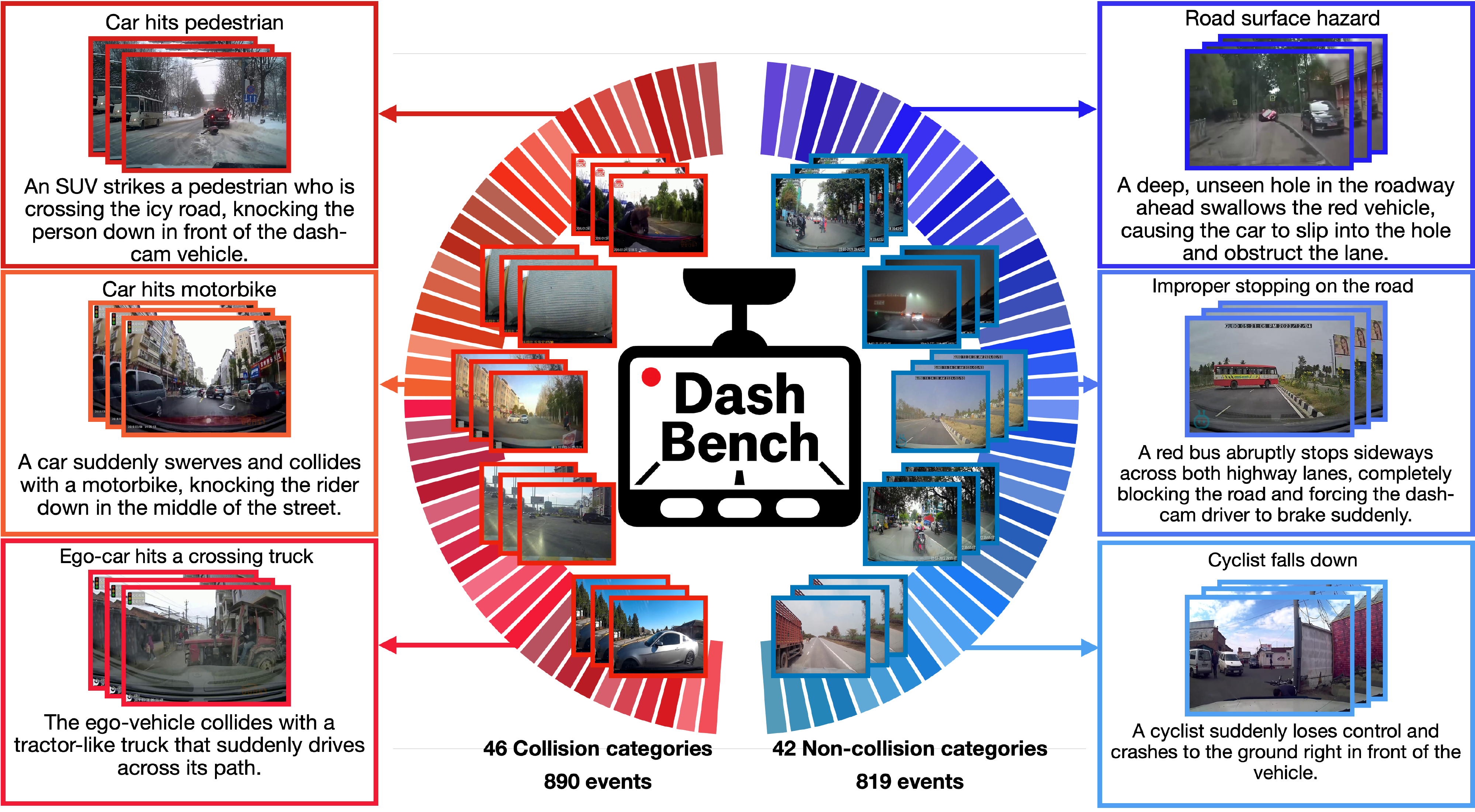}
    \vspace{6pt}
    \caption{Overview of \textbf{DashBench}. Our dataset spans 88 fine-grained categories to cover diverse corner case scenarios. Sample keyframes of corner case scenarios as well as their categories and annotations are shown.}
    \label{fig:overview}
\end{figure}

\section{Introduction}

Autonomous driving systems (ADS) face a fundamental transparency barrier: 
the learned models that process continuous dashcam video streams and translate 
perception into planning decisions \cite{DBLP:journals/pami/ChenWCJGL24} 
generate no real-time account of their reasoning, yet real-time natural-language explanations of individual maneuvers are 
increasingly recognised as an operational necessity: they support passenger 
awareness, enable operator oversight, and are a decisive factor in how 
drivers calibrate their reliance on the system 
\cite{dong2023did, zhang2024critical}.
Video Large Language Models (Video LLMs), multimodal models capable of reasoning over video inputs and generating grounded language descriptions have emerged as the leading 
candidate to close this interpretability gap in autonomous driving (AD). Video LLMs have been adapted for driving-specific scene comprehension, safety-critical event understanding, and language-grounded planning \cite{DBLP:conf/aaai/ChenLM0025, DBLP:journals/corr/abs-2412-07689}; leading commercial ADS have begun incorporating them for real-time narration \cite{waymo2024emma, wayve2024lingo2, kodiak2024llms}. Whether Video LLMs can fulfil this promise under the most demanding 
road conditions which are rare, safety-critical events far outside routine 
traffic which are referred to as \textit{corner cases} \cite{DBLP:conf/iccvw/BogdollBHBSFZ21, DBLP:journals/corr/abs-2102-05897}, is a question made especially urgent by a possible failure mode of Video LLMs. These models 
do not merely underperform on hard scenes, they can hallucinate and systematically omit important 
details from their descriptions 
\cite{DBLP:conf/cvpr/LiIF25}, producing active 
misinformation precisely where accurate perception matters most \cite{DBLP:conf/iccv/RawalSHSG25}.

Although there have been several works probing Video LLMs' scene understanding in corner cases for AD, they present several distinguishable limitations: \emph{(1) Insufficient coverage of corner case diversity}. Existing benchmarks in this domain \cite{DBLP:journals/corr/abs-2504-14526, DBLP:conf/wacv/0023LZLL0HTZLYL25, DBLP:journals/corr/abs-2501-04003} restrict themselves either to collision-related events \cite{DBLP:journals/corr/abs-2504-14526} or to object-level corner case detection \cite{DBLP:conf/wacv/0023LZLL0HTZLYL25}, leaving scenario-level non-collision hazards largely untested \cite{DBLP:journals/corr/abs-2102-05897}. \emph{(2) Limitations of experimental design and evaluation metrics}. DVBench \cite{DBLP:journals/corr/abs-2504-14526}, the benchmark most closely related to ours, relies solely on a multiple-choice question (MCQ) format, narrowing evaluation to option-picking and providing limited evidence of the model's genuine perception and explanatory reasoning \cite{DBLP:conf/iccv/KoLCCPK23}; subsequent work has highlighted the need for nuanced automatic judges to evaluate free-text responses in AD scenarios \cite{DBLP:journals/corr/abs-2501-04003}. \emph{(3) Lack of evaluation of driving-specialized Video LLMs on corner case scenarios}. Existing benchmarks that do evaluate models on driving corner cases assess general-purpose models exclusively and do not include driving-specialist Video LLMs \cite{DBLP:conf/eccv/MaCSPX24, DBLP:journals/corr/abs-2412-07689} that are explicitly designed and fine-tuned for driving video understanding. Whether such driving-specialized models close the corner-case reasoning gap over general-purpose Video LLMs remains an unanswered and practically important question, especially for guiding the development of road-ready Video LLMs.

We present \textbf{DashBench}, the first benchmark to not only systematically stress-test Video LLMs on corner case scenarios on the road under a multi-prompt evaluation protocol but also investigate the challenges presented by our task through anomaly-localized experiments and hallucination mitigation efforts in order to gain deeper insights. Our contributions are: 
\begin{enumerate}
     \item We curate, annotate, and process a dataset of \textbf{1,709} dashcam-video clips spanning \textbf{88} categories explicitly balanced across collision-related and non-collision hazardous events. As shown in Figure~\ref{fig:overview}, our dataset covers a broad array of scenario-level events.

     \item We evaluate \textbf{25} Video LLMs including SOTA closed-source models, open-source models, and driving-specialist Video LLMs fine-tuned on AD datasets, and find a consistent capability gap: even the highest-scoring model reaches only 46.52\% accuracy, well below any threshold that safety-critical deployment would demand. Notably, model design such as reasoning and driving-specialist fine-tuning yields no clear, consistent advantage over general-purpose models on corner-case scenarios.

     \item We document systematic failure modes through qualitative analysis of model outputs which are \textbf{corner-case blindness} (the model produces a fluent, confident description that entirely ignores the hazard) and \textbf{road agent/action hallucination} (non-existent vehicles, objects, or actions of other agents are fabricated). As shown in Figures~\ref{fig:pipeline}, \ref{fig:CCW}, and \ref{fig:mitigation}, these failures are not isolated errors but structural patterns that recur across model families.

     \item We demonstrate through extensive experiments that neither varying prompt formulations nor providing temporally localized corner case windows (isolating the frames where the hazard occurs) substantially improves corner-case reasoning, with performance gains remaining well below what safety-critical driving applications require. Additionally, applying existing hallucination mitigation techniques yields no consistent gains on safety-critical scenarios. Together, these findings suggest that closing the corner-case gap requires fundamentally new approaches.
\end{enumerate}

\section{Related Works}

\subsection{Open-Ended QA with Video LLMs}
Closed-set, multiple-choice video-QA formats \cite{DBLP:conf/nips/MangalamAM23} are susceptible to inflating apparent model capability by rewarding answer-option 
pattern matching \cite{DBLP:journals/corr/abs-2410-07752}, motivating a shift to open-ended, free-text evaluation. Video-MME \cite{DBLP:conf/cvpr/FuDLLRZWZSZCLLZ25} evaluates 
Video LLMs across diverse domains and task types and shows that 
state-of-the-art models fall substantially below human-level 
performance, while TemporalBench \cite{DBLP:journals/corr/abs-2410-10818} 
specifically targets fine-grained temporal understanding and finds 
similarly large gaps. Beyond format, evaluation is sensitive to prompt design: the 
same question phrased differently substantially affects output 
quality and can reverse rankings between models 
\cite{DBLP:journals/corr/abs-2512-17875}. Reliable capability assessment of Video LLMs therefore demands multiple prompt formulations across free-text, multi-frame conditions.

\subsection{Benchmarking Video LLMs for Corner Cases in AD}
Early benchmarks for driving video understanding 
evaluate Video LLMs exclusively on routine traffic conditions 
\cite{DBLP:conf/eccv/MarcuCHKHCNBKSAS24}, without systematic 
coverage of safety-critical events. This gap was formalized by \cite{DBLP:conf/wacv/0023LZLL0HTZLYL25}, 
establishing CODA-LM as the first benchmark to probe VLMs 
on corner cases, though limited to static, object-level 
anomalies. DVBench 
\cite{DBLP:journals/corr/abs-2504-14526} evaluated 14 
Video LLMs on 10K crash and near-crash MCQs, finding 
that no model exceeded 40\% accuracy, yet the closed-set 
format could not provide deeper insights on whether failures stemmed from 
perceptual blindness, reasoning errors, or output 
formatting. CrashSight 
\cite{DBLP:journals/corr/abs-2604-08457} broadens this 
further to roadside-camera footage with phase-aware 
reasoning tasks. Despite these advances, all existing 
corner-case benchmarks remain narrow, covering either 
object-level anomalies or crash-only events, and none systematically characterize the types of 
perceptual errors that drive underperformance. No benchmark to date provides 
diverse scenario-level coverage with open-ended, 
multi-prompt free-text scoring.

\begin{figure}[t]
    \centering
    \includegraphics[width=0.8\linewidth]{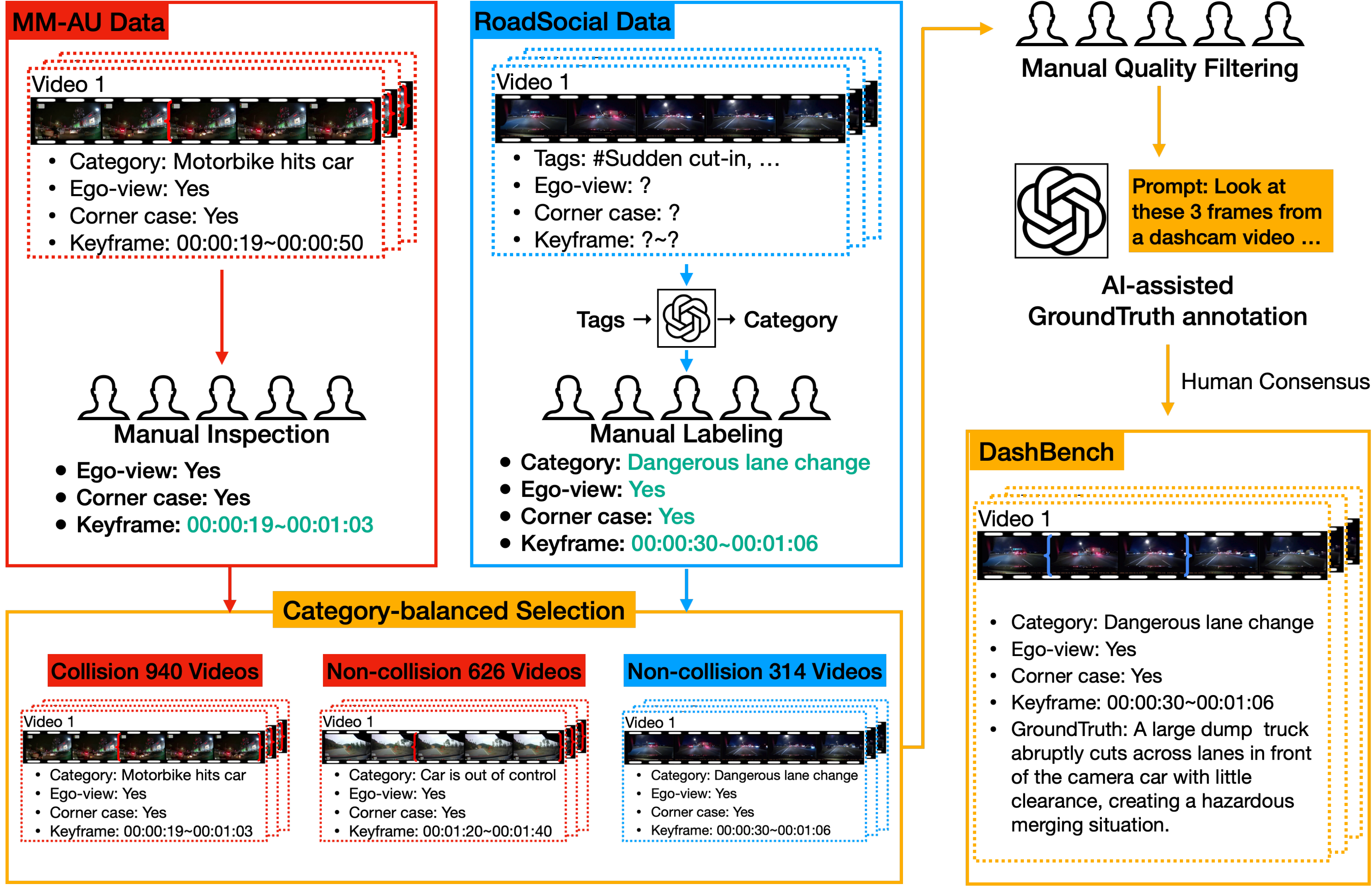}
    \vspace{9pt}
    \caption{DashBench construction pipeline. MM-AU and RoadSocial videos pass through source-specific filtering and annotation steps before a category-balanced merge, quality check, and AI-assisted ground truth annotation produce the final 1,709-clip benchmark.}
    \label{fig:dataset_pipeline}
\end{figure}

\section{DashBench}

We present \textbf{DashBench}, a benchmark of 1,709 ego-view dashcam
video clips designed to evaluate Video LLMs on scenario-level corner
cases in autonomous driving, spanning 88 corner case categories drawn
from two complementary sources: MM-AU~\cite{DBLP:conf/cvpr/FangLZXYLXC24} and RoadSocial~\cite{DBLP:conf/cvpr/ParikhRTGS25}, with a deliberate balance between collision and non-collision scenarios. Unless otherwise stated, 16 frames are uniformly sampled per clip as a standardized multi-frame evaluation protocol; benchmark construction is described thoroughly in this section, and the experimental setup is presented in Section~\ref{sec:experiments}.

\subsection{Motivation}                         

Driving corner cases are broadly defined as rare and unexpected situations that deviate from normal traffic behavior, challenging the reliability of autonomous perception systems~\cite{DBLP:conf/ivs/BreitensteinTLF20, DBLP:journals/corr/abs-2102-05897, DBLP:conf/iccvw/BogdollBHBSFZ21}. Following the scenario-level corner case taxonomy in \cite{DBLP:conf/ivs/BreitensteinTLF20}, DashBench targets scenario-level corner-cases which are well-suited for Video LLM evaluation since recognition of these events requires temporal reasoning over a video sequence. Within this level, the taxonomy separates anomalous, novel, and risky scenarios primarily by whether they carry collision potential. Building on this, our distinct dataset design splits clips by collision outcome (which is whether physical impact occurred). Collision clips capture scenarios in which the hazard resulted in actual impact; non-collision clips span the remaining range, from averted dangers and near-misses to unusual but safe road events.

We build DashBench's primary video corpus from \textbf{MM-AU}, the most fine-grained publicly available ego-view dashcam accident benchmark with 57 annotated categories and keyframe annotations. However, dashcam footage is predominantly collected for collision-related accident documentation~\cite{DBLP:conf/accv/ChanCXS16} and datasets curated from such footage therefore systematically over-represent collision events. Out of MM-AU's 57 categories, 46 describe direct collision events, yielding 9,142 collision samples against only 626 non-collision samples  which is a ratio of approximately 14.6:1 and a collision share of 93.6\% of all annotated clips. Underrepresented non-collision categories include vehicle loss of control, traffic rule violations without impact, and various road surface and environmental hazards, even though these are scenarios which are safety-critical as well. Therefore, a benchmark constructed solely from MM-AU would cover almost exclusively scenarios in which collision potential resulted in actual impact, leaving other real-world safety-critical events that do not end in collision absent from evaluation.

Therefore, to achieve broad scenario-level coverage, we augment the samples we will select from MM-AU with
non-collision corner case clips from \textbf{RoadSocial}, a
large-scale social-media road event dataset spanning varied geographies,
camera perspectives, and road event types. This dataset captures a wide spectrum of non-collision events absent from typical accident-focused datasets, and its video-level descriptive tags provide a principled basis for mapping clips to standardized corner case categories. Though we do not claim exhaustive coverage of all possible corner cases as no finite collection of videos can enumerate all possible real-world road hazards~\cite{DBLP:journals/corr/abs-2102-05897}, we believe that our constructed dataset provides the most wide-ranging and balanced ego-view collection yet assembled for scenario-level corner case evaluation, spanning 88 different categories.

\subsection{Dataset Composition}

\label{sec:composition}

As established in our motivation, non-collision corner
cases are severely underrepresented in existing ego-view dashcam datasets;
to address this gap and achieve balanced scenario-level coverage,
DashBench is composed of two sources. From MM-AU, we first retain all 626 non-collision clips spanning 11 categories (including vehicle loss of control and component failures); 
although MM-AU already provides ego-view footage and keyframe annotations, each clip still undergoes manual inspection to confirm that the corner case is visually identifiable and correctly categorized, and to refine keyframe boundaries where needed. To extend coverage beyond these 11 categories, we additionally draw on RoadSocial, comprising of 13.2K social-media road event videos across different camera viewpoints, each associated with descriptive tags out of a vocabulary of 21,325 video-level tags. Filtering and preparing these clips required three steps: an LLM-assisted pipeline mapping video-level tags to candidate corner case categories to identify relevant clips, a check to ensure clips exhibit ego-view dashcam perspective, and manual verification that a corner case is visually present, with annotators additionally labeling corner case keyframes (frame at the start of the corner case, the frame at which it is most observable, and the frame marking its end). This selection and annotation process was carried out by five annotators over four months, and the complete protocol is described in Figure~\ref{fig:dataset_pipeline}. Through this pipeline, RoadSocial yields 314 clips covering 31 new corner case categories.

For the collision-side subset, we sample from MM-AU's 9,142 collision
clips across all 46 collision categories, targeting a category-balanced
selection matching the size of the constructed non-collision subset. MM-AU provides keyframe annotations for the corner cases in all clips, but as previously mentioned, we manually refine keyframe boundaries where needed (cases where the keyframe does not exactly show the start, end, or the most observable frame of the corner case event).

To ensure that all clips are evaluable under the 16-frame uniform
sampling protocol, we apply two quality criteria across all candidate
clips from both sources. First, we ensure that the corner case must be self-contained (not requiring additional context outside of the images, e.g. sound/narration). Second, the corner case must be unambiguously identifiable from the 16 uniformly sampled frames. The final \textbf{DashBench} benchmark comprises \textbf{1,709 clips} across \textbf{88 corner case categories}. The average length for all clips in our benchmark is 7.70s while the average length of the corner case window for all clips is 3.47s.  The full category list along with sample counts are provided in Table~\ref{tab:categorization} in the Appendix.

\begin{wrapfigure}{r}{0.50\textwidth}
  \vspace{-\intextsep}
  \centering
  \includegraphics[width=\linewidth]{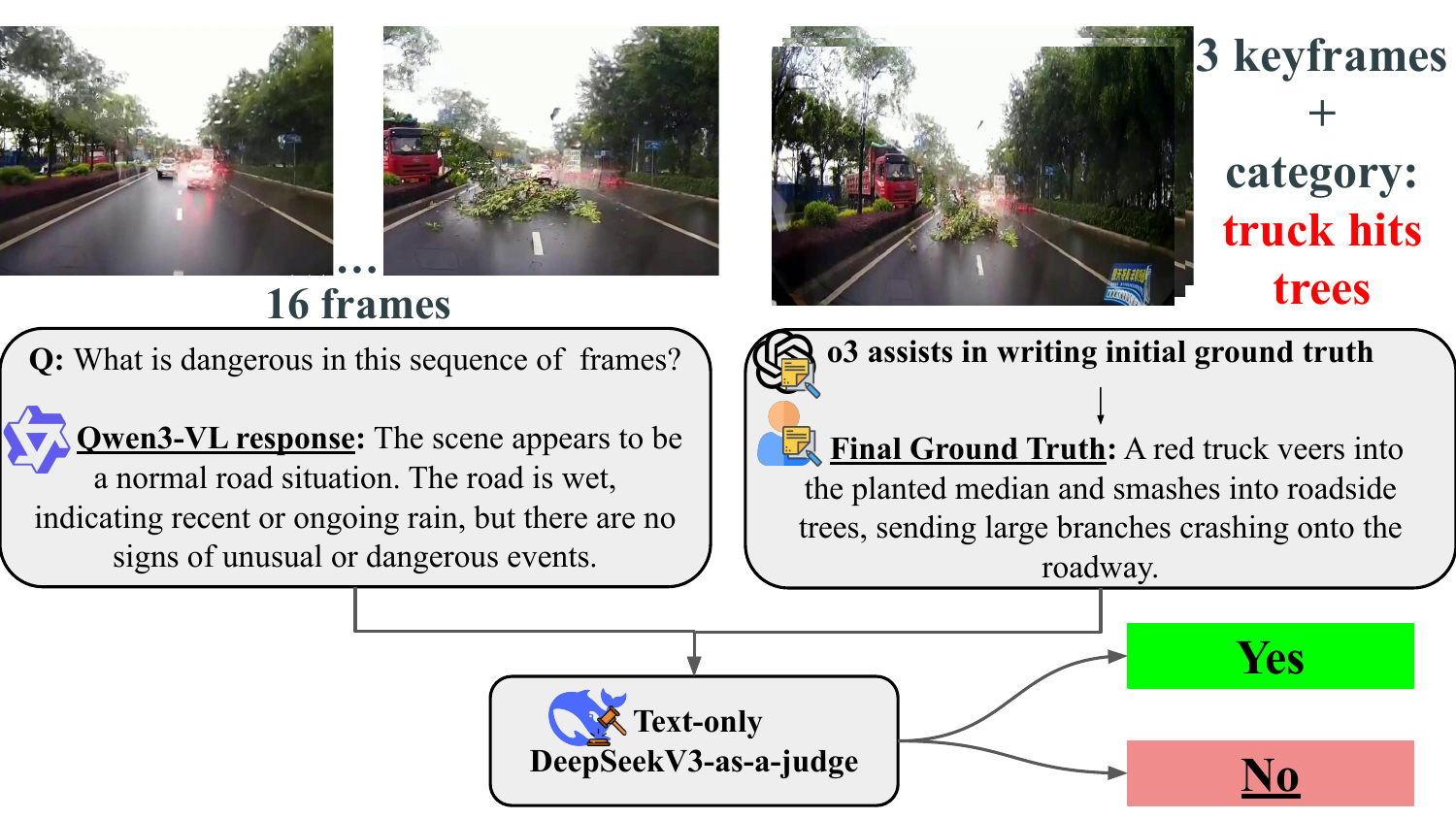}
  \vspace{3pt}
  \caption{Illustration of how we obtain a model's response and ground truth description and pass them to \texttt{\deepseekvthree}-as-a-judge.}
  \label{fig:pipeline}
  \vspace{-\intextsep}   %
\end{wrapfigure}

\subsection{AI-Assisted Ground Truth Annotation} 
Writing detailed scene descriptions for thousands of clips from the ground up is labour-intensive; we therefore leverage a strong thinking VLM, OpenAI's \oThree model which can reason with images in its chain of thought \cite{o3ando4mini}, to assist us in writing ground truth annotations. In each sample's prompt, we provide three keyframes (video start frame, corner case occurrence frame, and video end frame) and pass the corner case category as a cue to guide \oThree's annotation. Three independent annotators refine each generated model caption to a consensus final label, ensuring reliable ground-truth quality. The detailed prompt and annotation process can be found in Appendix \ref{appsec:prompts} and \ref{app:gt_annotation}, respectively.

\section{Experiments}
\label{sec:experiments}

\subsection{Experimental Setups}

DashBench is an open-ended, free-text response benchmark. While early video-QA datasets often relied on fixed multiple-choice formats, recent studies highlight that these multiple-choice tasks allow models to exploit shortcuts, revealing little about genuine multimodal reasoning or visual grounding \cite{DBLP:conf/iccv/KoLCCPK23, DBLP:conf/nips/LiuMQWSC24}. As a result, recent benchmark efforts have shifted toward open-ended evaluation to properly stress-test Video LLMs' understanding of time-sensitive and knowledge-intensive visual events \cite{DBLP:conf/cvpr/DiX24, DBLP:journals/corr/abs-2601-02536}. Figure \ref{fig:pipeline} shows the concise pipeline of experiments in our dataset, from how we obtain both a model's response and a ground truth description to then subsequently pass them to \deepseekvthree-as-a-judge for evaluation.

\subsubsection{Prompts}

For each sample, we query the Video LLM with five prompts:

\begin{enumerate}[label=\textbf{(Q\arabic*)},  %
                 labelsep=0.6em,
                 align=left,
                 leftmargin=*]

    \item[\textbf{(Q1)}]
    \emph{What is unusual in this sequence of frames?}

    \item[\textbf{(Q2)}]
    \emph{What is dangerous in this sequence of frames?}

    \item[\textbf{(Q3)}]
    \emph{Describe this sequence of frames in detail.}

    \item[\textbf{(Q4)}]
    Asks the model to return a single JSON object with two fields:
    \begin{enumerate}
        \item (Q4\textsubscript{TF}) -- \texttt{true} if there is an unusual/dangerous event and \texttt{false} if it is a normal road situation
        \item (Q4\textsubscript{R}) -- free-text reasoning that justifies the answer in (Q4\textsubscript{TF})
    \end{enumerate}

    \item[\textbf{(Q5)}]
    First gives a definition of a corner case scenario along with some examples. Then, asks the model to identify if there is any corner case and to describe it.
\end{enumerate}

\textbf{Q1–Q2} are designed as explicit questions to force the model to search for anomalies, using both \emph{unusual} and \emph{dangerous} to cover diverse wording that can be used to describe corner cases. \textbf{Q3} takes a different approach by asking for a general scene description with no explicit mention of danger, anomalies, or corner cases. We consider this a critical test since a Video LLM that truly understands the unfolding scene should identify the evident corner case naturally, without any textual hint to guide it. \textbf{Q4} serves as an integrated detection-and-explanation prompt where we jointly check two aspects: (i) \textbf{Q4\textsubscript{TF}}, the binary question format is designed to check whether the model might be sensitive to the wording in the prompt, and (ii) in \textbf{Q4\textsubscript{R}} we ask the Video LLM to give its reasoning as to what factors in the images display a routine traffic scene or corner case on the road. This "reasoning" test examines the Video LLM's visual grounding capability and whether it guesses in \textbf{Q4\textsubscript{TF}} due to the prompt's nature. Since the format of \textbf{Q4} requires the model to return JSON outputs, we could not provide \textbf{Q4} results on several models since they cannot do this consistently. Lastly, \textbf{Q5} introduces a definition-guided setting: we provide the model with a description of what a corner case is as well as some examples, then direct it to identify and describe any such event in the clip. Video-QA benchmarks generally avoid including in-context examples directly in the prompt \cite{DBLP:conf/nips/SalehiPKK0HF24, DBLP:conf/nips/LiuMQWSC24, DBLP:conf/eccv/MarcuCHKHCNBKSAS24}, as video inputs already consume a large context budget and make adding examples computationally expensive. \textbf{Q5} therefore stays within this convention by providing contextual framing through a short textual definition and examples, without adding concrete video instances to the prompt. We also use a system prompt to encourage the model to answer concisely as a road assistant. Detailed prompt templates are provided in Appendix~\ref{appsec:prompts}

\subsubsection{Models}

We perform a comprehensive evaluation of 25 Video LLMs: 2 SOTA closed-source models, \gptfive~\cite{DBLP:journals/corr/abs-2601-03267} and \geminiflash~\cite{pichai2025new}, 16 general-purpose open-source models including \qwenvl~\cite{DBLP:journals/corr/abs-2511-21631} (\texttt{8B}, \texttt{32B}, and \texttt{235B} variants), \internvlthreefive~\cite{DBLP:journals/corr/abs-2508-18265} (\texttt{8B}, \texttt{38B}, and \texttt{241B} variants), \ovistwofive~\cite{DBLP:journals/corr/abs-2508-11737}, \keyevl~\cite{DBLP:journals/corr/abs-2509-01563}, \minicpmfourfive~\cite{DBLP:journals/corr/abs-2509-18154}, \glmfouronevthinking~\cite{DBLP:journals/corr/abs-2507-01006}, \llavanextvideo~\cite{DBLP:journals/tmlr/ZhangWLLMLL25}, \tarsiertwo~\cite{DBLP:journals/corr/abs-2501-07888}, \vitaonefive~\cite{DBLP:journals/corr/abs-2501-01957}, \internlmxcomposertwo~\cite{DBLP:journals/corr/abs-2407-03320}, \internvideo ~\cite{DBLP:journals/corr/abs-2501-12386} and \llavaonevision~\cite{DBLP:journals/corr/abs-2509-23661}, as well as a group of 7 Video LLMs, which we will refer to as \emph{specialists}, that have been fine-tuned on different established AD datasets. These include \dolphins~\cite{DBLP:conf/eccv/MaCSPX24}, fine-tuned on the BDD-X dataset~\cite{DBLP:conf/eccv/KimRDCA18}, \drivemm~\cite{DBLP:journals/corr/abs-2412-07689} which was fine-tuned on 6 separate AD datasets to promote versatility in handling different AD tasks, two domain-adapted versions of \miniinternvl~\cite{DBLP:journals/corr/abs-2410-16261} fine-tuned on the DriveLM dataset~\cite{DBLP:conf/eccv/SimaRCCZXBLGL24} and the BDD dataset~\cite{DBLP:conf/cvpr/YuCWXCLMD20} respectively, \drivelmm~\cite{DBLP:conf/iros/IshaqLMTTDALKCLAK25} which incorporates step-by-step chain-of-thought reasoning adapted to driving scene understanding, \crashchat~\cite{DBLP:journals/corr/abs-2512-18878} fine-tuned for multitask traffic crash video analysis, and \llavaroadsocial, a 7B model fine-tuned on the training split of the RoadSocial dataset~\cite{DBLP:conf/cvpr/ParikhRTGS25}. See Appendix~\ref{appsec:models} for details on model access and deployment.

\subsubsection{Evaluation} 

\begin{table*}[!t]   
\centering              
\footnotesize
\setlength{\tabcolsep}{3pt}
\renewcommand{\arraystretch}{1.1}
\begin{tabular}{l|ccccc|ccc|c}
\hline
\multirow{2}{*}{\textbf{Model}} &
\multirow{2}{*}{\centering\textbf{Q1}} &
\multirow{2}{*}{\centering\textbf{Q2}} &
\multirow{2}{*}{\centering\textbf{Q3}} &
\multirow{2}{*}{\centering\textbf{Q4\textsubscript{R}}} &
\multirow{2}{*}{\centering\textbf{Q5}} &
\multirow{2}{*}{\parbox{1cm}{\centering\textbf{Min}\\\textbf{Agree}}} &
\multirow{2}{*}{\parbox{1cm}{\centering\textbf{Max}\\\textbf{Agree}}} &
\multirow{2}{*}{\parbox{1cm}{\centering\textbf{Avg}\\\textbf{Agree}}} &
\multirow{2}{*}{\centering\textbf{Q4\textsubscript{TF}}} \\
& & & & & & & & & \\ \hline

\multicolumn{10}{l}{\textbf{Closed-source}} \\
\hline
\gptfive              & \underline{19.95} & \textbf{36.92} & \textbf{31.19} & \textbf{46.52} & \textbf{39.15} & 19.95 & 46.52 & 34.75 & 97.83 \\
\geminiflash          & \textbf{32.36} & \underline{35.28} & \underline{29.84} & \underline{39.58} & \underline{34.99} & 29.84 & 39.58 & 34.41 & 94.85 \\

\hline
\multicolumn{10}{l}{\textbf{Open-source}} \\
\hline
\qwenvl \texttt{(235B)}        & 18.72 & \textbf{20.19} & \textbf{16.79} & \textbf{25.92} & \textbf{21.71} & 16.79 & 25.92 & 20.67 & 72.03 \\
\qwenvl \texttt{(8B)}          & 18.90 & 15.51 & \underline{14.45} & \underline{22.00} & \underline{21.53} & 14.45 & 22.00 & 18.48 & 74.96 \\
\qwenvl \texttt{(32B)}         & 17.20 & \underline{18.90} & 12.87 & 18.78 & 18.31 & 12.87 & 18.90 & 17.21 & 71.04 \\
\ovistwofive \texttt{(9B)}     & \textbf{19.68} & 15.41 & 7.77  & 18.72 & 20.55 & 7.77  & 20.55 & 16.43 & 64.47 \\
\internvlthreefive \texttt{(241B)} & \underline{19.31} & 12.81 & 11.47 & 19.13 & 17.61 & 11.47 & 19.31 & 16.07 & 76.36 \\
\internvlthreefive \texttt{(38B)} & 12.64 & 13.22 & 8.66  & 16.44 & 12.93 & 8.66  & 16.44 & 12.78 & 77.65 \\
\tarsiertwo \texttt{(7B)}                 & 13.28 & 13.11 & 11.70 & --    & 12.35 & 11.70 & 13.28 & 12.61 & --    \\
\minicpmfourfive \texttt{(8B)}            & 13.34 & 11.18 & 6.38  & 17.27 & 13.53 & 6.38  & 17.27 & 12.34 & 87.53 \\
\keyevl \texttt{(8B)}          & 12.40 & 11.01 & 5.03  & 13.13 & 15.69 & 5.03  & 15.69 & 11.45 & 80.09 \\
\internvlthreefive \texttt{(8B)} & 4.27  & 7.67 & 6.55  & 12.17 & 12.52 & 4.27  & 12.52 & 8.64 & 78.17 \\
\internvideo \texttt{(8B)}                & 4.86  & 9.71  & 4.15  & 9.95  & 10.12 & 4.15  & 10.12 & 7.76  & 54.00 \\
\internlmxcomposertwo \texttt{(7B)}       & 6.26  & 9.89  & 4.92  & 9.07  & 8.54  & 4.92  & 9.89  & 7.74  & 54.44 \\
\glmfouronevthinking \texttt{(9B)}        & 7.85  & 5.50  & 2.05  & 6.32  & 9.91  & 2.05  & 9.91  & 6.33  & 35.34 \\
\vitaonefive \texttt{(8B)}                & 7.31  & 3.63  & 3.16  & 6.03  & 7.08  & 3.16  & 7.31  & 5.44  & 61.32 \\
\llavaonevision \texttt{(8B)}             & 3.28  & 3.10  & 1.46  & 7.14  & 6.90  & 1.46  & 7.14  & 4.38  & 54.13 \\
\llavanextvideo \texttt{(7B)}  & 1.11  & 3.16  & 0.18  & 2.00  & 1.40  & 0.18  & 3.16  & 1.57  & 67.12 \\

\hline
\multicolumn{10}{l}{\textbf{Specialists}} \\
\hline
\llavaroadsocial \texttt{(7B)}            & \underline{6.09} & \textbf{14.86} & \textbf{11.59} & -- & \textbf{7.49} & 6.09 & 14.86 & 10.01 & -- \\
\crashchat \texttt{(7B)}                  & \textbf{6.79} & \underline{8.84}  & \underline{4.27}  & -- & \underline{3.74} & 3.74 & 8.84  & 5.91 & -- \\
\drivelmm \texttt{(8B)}                   & 3.92 & 2.28  & 1.11  & -- & 3.74 & 1.11 & 3.92  & 2.76  & -- \\
\drivemm \texttt{(8B)}                    & 1.70 & 1.76  & 0.06  & -- & 2.17 & 0.06 & 2.17  & 1.42  & -- \\
\miniinternvlbdd \texttt{(4B)}            & 2.28 & 0.64  & 0.00  & -- & 1.46 & 0.00 & 2.28  & 1.10  & -- \\
\miniinternvldrivelm \texttt{(4B)}        & 0.82 & 0.35  & 0.12  & -- & 2.17 & 0.12 & 2.17  & 0.87  & -- \\
\dolphins \texttt{(7B)}                   & 0.12 & 1.05  & 0.00  & -- & 0.12 & 0.00 & 1.05  & 0.32  & -- \\

\hline
\end{tabular}
\vspace{9pt}
\caption{Video LLMs' accuracy (\%) on all corner-case driving scenarios in \textbf{DashBench}.  "Minimum/Maximum/Average Agreement" are computed over Q1, Q2, Q3, Q4\textsubscript{R}, and Q5, and models are ordered based on Average Agreement in their own family. Q4\textsubscript{TF} is reported separately because it simply measures event detection rather than grounded explanation. Bold numbers mark the best value in each Q1–Q5 column within a model family; underlined numbers mark the second-best. \texttt{--} denotes cases where models could not consistently respond to Q4 in a structured format.}
\label{tab:dashbench_results}
\vspace{-\intextsep}
\end{table*}

To reliably evaluate the alignment between Video LLM responses and ground truth annotations, we employ the LLM-as-judge evaluation protocol \cite{DBLP:conf/nips/ZhengC00WZL0LXZ23}. This has become the standard practice for open-ended benchmarks, with recent work employing GPT 
variants \cite{DBLP:conf/eccv/MarcuCHKHCNBKSAS24, DBLP:conf/wacv/0023LZLL0HTZLYL25}, \deepseekvthree \cite{DBLP:conf/emnlp/RolshovenRBHBSN25}, and \llama \cite{DBLP:journals/corr/abs-2412-15524} as evaluators. In our setup, the LLM-as-judge assesses Q1--Q3 and Q4\textsubscript{R} using a structured binary judgment prompt that asks whether each model response captures the same corner case as the ground truth in terms of its cause, the agents responsible, and event timing. This criterion counts only responses receiving a \emph{Yes} verdict toward the accuracy score, ensuring a metric that reflects semantic alignment, while Q4\textsubscript{TF} is reported separately as an independent binary detection score. We validate LLM-as-judge selection by randomly sampling 100 responses of several representative Video LLMs and comparing each candidate judge's verdicts on those responses to actual human judgments using Cohen's kappa ($\kappa$)~\cite{cohen1960coefficient} as the inter-rater agreement measure. As shown in Table~\ref{tab:judge_kappa}, \deepseekvthree achieves the highest agreement with human judgments ($\kappa = \mathbf{0.96}$) among all candidates tested, including \llama~\cite{meta2024introducing} and a multi-LLM ensemble based on \cite{DBLP:journals/corr/abs-2404-18796} (the details for this ensemble setup are explained in Appendix~\ref{appsec:evaluation}), so we select \deepseekvthree as our LLM-as-judge. As an open-source model from a different family than the evaluated Video LLMs, \deepseekvthree additionally supports reproducible evaluation and avoids circular self-evaluation bias \cite{DBLP:conf/nips/PanicksseryBF24}. Furthermore, to verify that our evaluation scheme is not sensitive to stochastic variation, we run the \deepseekvthree-as-a-judge evaluation five times on a representative set of model responses; the mean accuracy and standard deviation across all prompts and models confirm consistent evaluation (std $<$ 1\% in all cases). The complete judge prompt is shown in Appendix~\ref{appsec:prompts}, while the details for the judging experiments including multi-run stability results are saved in Table~\ref{tab:judge_variance} in the Appendix. 

\begin{table}[t]
    \centering
    \footnotesize
    \setlength{\tabcolsep}{2.8pt}
    \renewcommand{\arraystretch}{1.12}

    \begin{tabular}{l|ccccc|c}
    \hline
    \multirow{2}{*}{Judge} &
    \multirow{2}{*}{\parbox{1cm}{\centering \gptfive \\ $\kappa$}} &
    \multirow{2}{*}{\parbox{1.3cm}{\centering \texttt{Gemini-3} \\ \texttt{-Flash} $\kappa$}} &
    \multirow{2}{*}{\parbox{1.3cm}{\centering \qwenvl \\ \texttt{(8B)}$\kappa$}} &
    \multirow{2}{*}{\parbox{1.8cm}{\centering \internvlthreefive \\ \texttt{(8B)} $\kappa$}} &
    \multirow{2}{*}{\parbox{1.3cm}{\centering \ovistwofive \\ \texttt{(9B)} $\kappa$}} &
    \multirow{2}{*}{\parbox{1cm}{\centering Avg \\ $\kappa$}} \\
    & & & & & & \\
    \hline

    \deepseekvthree & 0.98 & 0.93 & 0.93 & 0.95 & 1.00 & 0.96 \\
    \llama          & 0.90 & 0.92 & 0.76 & 0.81 & 0.85 & 0.85 \\
    Multi-LLM \cite{DBLP:journals/corr/abs-2404-18796}
                    & 0.63 & 0.57 & 0.44 & 0.52 & 0.54 & 0.54 \\

    \hline
    \end{tabular}

    \vspace{9pt}
    \caption{Cohen's $\kappa$ agreement between different judge setups and evaluated Video LLM responses.}
    \label{tab:judge_kappa}

\end{table}

\subsection{Results}

Table~\ref{tab:dashbench_results} reports the accuracy of all evaluated Video LLMs on the full dataset across all five prompts. No model surpasses $50\%$ on any individual prompt, and none exceeds $35\%$ on average, highlighting the fundamental difficulty of open-ended corner-case understanding. Furthermore, our extensive experiments reveal distinct patterns in how architectural choices and prompt design affect model capabilities. The results demonstrate clear differences based on model type, training approach, and prompt structure, with ADS deployment implications.

\paragraph{Corner-Case Understanding Remains Challenging Across All Model Types and Scale.} Among all models, \gptfive achieves the highest Q4\textsubscript{R} accuracy at a lacking $46.52\%$ and an average agreement of $34.75\%$, while \geminiflash follows closely at $34.41\%$ average. Though these numbers remain far below any practical safety threshold, the lead over the strongest open-source model \qwenvl (\texttt{235B}, $20.67\%$ avg), still exceeds $14\%$ in average agreement ($34.75\%$ vs.\ $20.67\%$), suggesting that our task remains challenging even as open-source general multimodal capabilities continue to improve. Notably, scaling model size within the same model family yields only modest and inconsistent gains. While scaling to extremely large open-source models (\texttt{235B+}) does improve performance, the gains remain surprisingly modest. For example, \qwenvl (\texttt{235B}) achieves only a $\sim$2\% higher average agreement than its \texttt{8B} counterpart ($20.67\%$ vs.\ $18.48\%$). Similarly, scaling \internvlthreefive from \texttt{8B} to \texttt{241B} improves average agreement by only $\sim$7.43\% ($8.64\%$ vs.\ $16.07\%$). These results suggest that simply increasing model scale alone is insufficient for robust understanding of rare and safety-critical driving corner cases.

\paragraph{Neither Reasoning nor Driving Specialization Solves Corner-Case Understanding.} Additionally, neither extended chain-of-thought reasoning nor domain-specific driving specialization appears sufficient for robust corner-case understanding. Reasoning-oriented models such as \ovistwofive, \keyevl, and \glmfouronevthinking show no clear advantage over similarly sized open-source Video LLMs, as \ovistwofive reaches $16.43\%$ average, behind the non-reasoning \qwenvl at $18.48\%$, while \glmfouronevthinking scores only $6.33\%$ despite its thinking mode. In particular, specialist driving Video LLMs achieve at most $10.01\%$ average agreement despite being explicitly fine-tuned on AD datasets and for AD tasks, with several models collapsing to near-random performance. This suggests that current specialist training strategies may encourage shortcut learning tied to benchmark-specific patterns rather than genuine understanding of complex driving events.

\paragraph{Video LLMs Depend Heavily on Linguistic Cueing Rather Than Robust Visual Understanding.} Variations in our prompts' wording reveal that model performance can be highly sensitive to semantic framing: the same scene queried under different linguistic formulations shows score swings of up to $26\%$  within a single model (e.g., $19.95\%$ on Q1 to $46.52\%$ on Q4\textsubscript{R} for \gptfive), a critical consideration for prompt design when deploying Video LLMs in driving scenarios. Additionally, through Q4\textsubscript{R} and Q4\textsubscript{TF}, we reveal the importance of \emph{free-text reasoning}: are Video LLMs' grounded explanations on successfully detecting a corner case valid? All evaluated models suffer a sharp drop in performance when asked to explain their decisions. On average, accuracy falls by \textbf{55 percentage points} between Q4\textsubscript{TF} and Q4\textsubscript{R}. This insight calls for the integration of additional modules that verify visual evidence before explaining. Most critically, Q3, which asks for a detailed temporal description of the video without explicitly cuing the model to look for corner cases, ranks as the lowest-scoring prompt across almost all evaluated models, regardless of model family or scale. Therefore, the consistent drop on Q3, which removes any implicit guidance on searching for corner cases in the scene, exposes that performance can be driven by what models are told to look for, not by what they can independently perceive.

\begin{figure}[t]
\centering
\includegraphics[width=\linewidth, height=0.42\textheight]{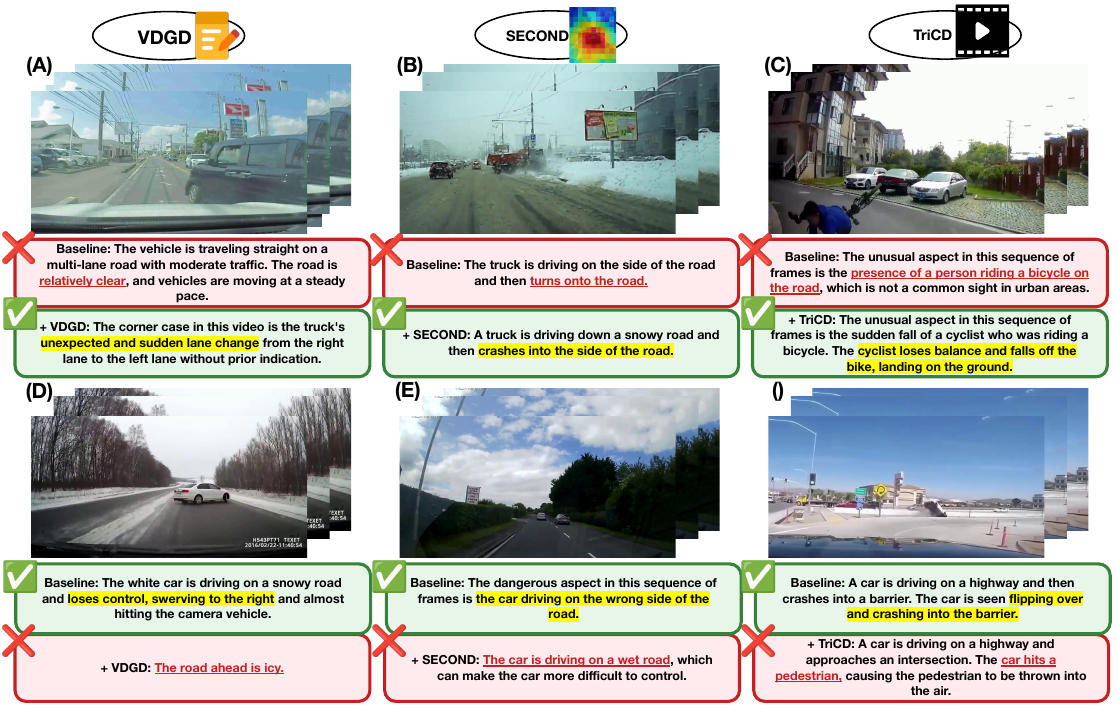}
\vspace{6pt}
\caption{
  \textbf{Samples of "flip" cases as a result of hallucination mitigation methods applied to the baseline model, \internvlthreefive \texttt{8B}.} \cmark~denotes a response that correctly identifies the corner case, while \xmark~denotes a response that fails to do so, either missing the hazard entirely or substituting a hallucinated description. The top row (A--C) shows recovery cases, whereas the bottom row (D--F) shows regression cases.
}
\label{fig:mitigation}
\end{figure}

\paragraph{Video LLMs Fail Both by Missing Hazards and Hallucinating Them.} Detailed qualitative analysis on our benchmark reveals the important issue on not only that Video LLMs fail but also ''how they fail''. Across many evaluated instances, Video LLMs simply struggle to identify the corner case in the scene (Figure \ref{fig:CCW}, \ref{fig:mitigation}, Appendix \ref{app:fail_ex}). Beyond failures in visual grounding, we observe that Video LLMs have a tendency to hallucinate nonexistent road agents, which are especially important in vision-language-action frameworks where imagined hazards will directly influence an ADS' road maneuvers. Especially in Q1 and Q2 results, these models might invent scenarios instead of claiming the absence of anything unusual or dangerous or even uncertainty (Figure \ref{fig:mitigation}, Appendix \ref{app:hallucinate_ex}). Though the resulting limited performance for all corner cases are most commonly related to failing to identify the corner case and hallucinating non-existent road agents  (Appendix \ref{app:fail_ex} and \ref{app:hallucinate_ex} respectively), observation of collision event examples specifically show that Video LLMs additionally hallucinate imaginary actions of road agents (Appendix~\ref{app:hallucinate_action_ex}) and fail to identify the actual collision, instead describing them as near-misses (Appendix \ref{app:coll_ex}). 

\subsubsection{Corner Case Window Experiments}

\begin{figure}[t]
\centering
\includegraphics[width=0.9\linewidth]{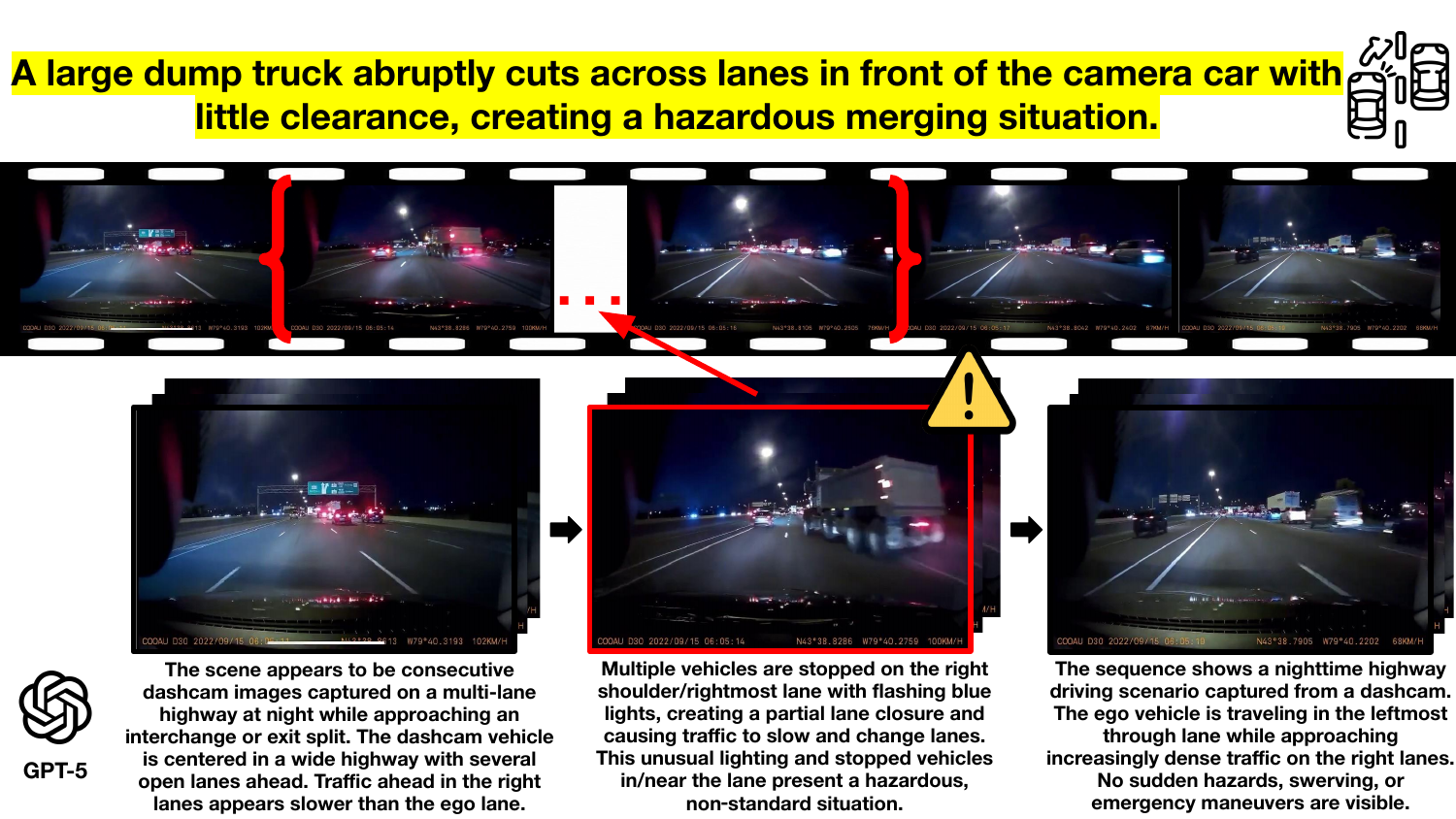}
\vspace{6pt}
\caption{
  \textbf{Illustration of a corner case window in a clip.} The red brackets mark the annotated 
  corner-case window in the clip. We show \gptfive's responses to different parts of the clip; \gptfive is able to explain the scenario well outside of the corner case window. However, despite using 16 frames and concentrating the visual input on 
  the critical event (\hl{a truck performing hazardous lane merging ahead}), \gptfive \textcolor{red!80!black}{fails to identify the dangerous maneuver}.
}
\label{fig:CCW}
\end{figure}

In our main evaluation, 16 frames are sampled uniformly across each 
dashcam clip. As illustrated in Figure~\ref{fig:CCW} (top), corner-case 
events occupy only a fraction of the clip on average, meaning the frame 
budget is shared between the hazardous segment and surrounding routine 
footage. The \emph{corner-case window} (CCW) 
experiment tests whether this is a limiting factor: rather than 
sampling from the full clip, we restrict the 16 frames to the 
annotated start-to-end span of the corner case only, keeping 
everything else such as prompts, models, and evaluation, identical. We apply this experiment to the \textbf{823} out of $1{,}709$ samples ($48.2\%$) in which the anomalous window covers less than half of the total clip duration. These are where the localization argument is strongest and uniformly sampling throughout the video is most likely to fail to sufficiently cover the anomalous segment within the frame budget.

As reported in Table~\ref{tab:ccw_results} as well as expanded results in Appendix Table \ref{tab:ccw_results_expanded}, restricting to the corner-case window improves average agreement, confirming that temporal
localization is broadly beneficial.
The magnitude of improvement seems to scale with model capability:
closed-source models gain $+3.79$ pp (\gptfive) and $+4.31$ pp (\geminiflash),
while open-source models improve by only $+0.81$ to $+1.98$ pp, suggesting that
stronger models are better positioned to exploit the concentrated event signal.
Notably, in Q1, both \gptfive ($-1.22$ pp) and
\qwenvl ($-4.37$ pp) regress on the narrow window.
Q1 asks what is \textit{unusual} about the scene, a judgment that inherently
requires comparative context from the surrounding normal driving footage;
excluding that context by narrowing to the event window alone removes the
reference frames against which the anomaly is assessed.
Taken together, these results suggest that while temporal event localization
meaningfully reduces the visual noise that hinders corner-case understanding,
the ceiling for open-source models still remains limited by capability rather than
by sampling strategy alone.

\begin{table}[!t]
\centering
\footnotesize
\setlength{\tabcolsep}{3pt}
\renewcommand{\arraystretch}{1.1}
\begin{tabular}{l|ccccc|c|c}
\hline
\multirow{2}{*}{\textbf{Model}} &
\multirow{2}{*}{\centering\textbf{Q1}} &
\multirow{2}{*}{\centering\textbf{Q2}} &
\multirow{2}{*}{\centering\textbf{Q3}} &
\multirow{2}{*}{\centering\textbf{Q4\textsubscript{R}}} &
\multirow{2}{*}{\centering\textbf{Q5}} &
\multirow{2}{*}{\parbox{1.2cm}{\centering\textbf{Avg}\\\textbf{Agree}}} &
\multirow{2}{*}{\centering\textbf{Q4\textsubscript{TF}}} \\
& & & & & & & \\ \hline
\multicolumn{8}{l}{\textbf{Closed-source}} \\
\hline
\gptfive{}
  & 20.90 & 36.09 & 29.28 & 47.39 & 38.40 & 34.41 & 96.96 \\
\rowcolor{gray!15}
\,+\,\textit{CCW}
  & $19.68_{\pminus{1.22}}$
  & $42.41_{\pplus{6.32}}$
  & $34.51_{\pplus{5.23}}$
  & $51.40_{\pplus{4.01}}$
  & $43.01_{\pplus{4.61}}$
  & $38.20_{\pplus{3.79}}$
  & $97.81_{\pplus{0.85}}$ \\
\geminiflash{}
  & 32.56 & 34.99 & 28.43 & 37.96 & 34.39 & 33.67 & 92.34 \\
\rowcolor{gray!15}
\,+\,\textit{CCW}
  & $34.63_{\pplus{2.07}}$
  & $39.37_{\pplus{4.38}}$
  & $30.13_{\pplus{1.70}}$
  & $45.44_{\pplus{7.48}}$
  & $40.34_{\pplus{5.95}}$
  & $37.98_{\pplus{4.31}}$
  & $94.29_{\pplus{1.95}}$ \\
\hline
\multicolumn{8}{l}{\textbf{Open-source}} \\
\hline
\qwenvl \texttt{(8B)}
  & 17.98 & 14.95 & 13.97 & 20.90 & 21.02 & 17.76 & 70.84 \\
\rowcolor{gray!15}
\,+\,\textit{CCW}
  & $13.61_{\pminus{4.37}}$
  & $18.35_{\pplus{3.40}}$
  & $15.19_{\pplus{1.22}}$
  & $22.11_{\pplus{1.21}}$
  & $23.57_{\pplus{2.55}}$
  & $18.57_{\pplus{0.81}}$
  & $73.39_{\pplus{2.55}}$ \\
\ovistwofive \texttt{(9B)}
  & 18.57 & 13.62 & \phantom{0}7.01 & 18.10 & 21.37 & 15.73 & 60.59 \\
\rowcolor{gray!15}
\,+\,\textit{CCW}
  & $20.79_{\pplus{2.22}}$
  & $17.30_{\pplus{3.68}}$
  & $\phantom{0}7.74_{\pplus{0.73}}$
  & $19.66_{\pplus{1.56}}$
  & $23.08_{\pplus{1.71}}$
  & $17.71_{\pplus{1.98}}$
  & $64.99_{\pplus{4.40}}$ \\
\internvlthreefive \texttt{(8B)}
  & \phantom{0}3.65 & \phantom{0}8.02 & \phantom{0}6.68 & 10.57 & 12.15 & \phantom{0}8.21 & 74.48 \\
\rowcolor{gray!15}
\,+\,\textit{CCW}
  & $\phantom{0}4.74_{\pplus{1.09}}$
  & $\phantom{0}9.11_{\pplus{1.09}}$
  & $\phantom{0}7.65_{\pplus{0.97}}$
  & $14.22_{\pplus{3.65}}$
  & $15.19_{\pplus{3.04}}$
  & $10.18_{\pplus{1.97}}$
  & $85.05_{\pplus{10.57}}$ \\
\hline
\end{tabular}
\vspace{9pt}
\caption{
    Accuracy (\%) of selected models on the CCW subset.
    Each model's full-video run (16 frames sampled across the full clip) is compared against
    its CCW run (16 frames sampled from the corner case window only).
    Subscripts show the change in score: \textcolor{green!55!black}{green} = improvement,
    \textcolor{red!80!black}{red} = regression.
}
\label{tab:ccw_results}
\end{table}

\subsubsection{Hallucination Mitigation On the Road}

Due to the difficulty of acquiring sufficient annotated corner-case examples and curating a class-balanced training set across 88 scenario categories, fine-tuning Video LLMs for corner-case robustness remains practically expensive. 
We therefore investigate recent training-free methods for multimodal LLMs that require 
no additional data or model updates, and whose implementations are publicly 
available. Specifically, we focus on \textit{training-free decoding}, which mitigate hallucinations at inference time by contrasting the model’s normal predictions with degraded or perturbed visual references, thereby amplifying visually grounded tokens and suppressing language-driven priors.

VDGD ~\cite{DBLP:conf/iclr/GhoshEKTNJM25} grounds generation using a detailed visual description and KL-based decoding. SECOND~\cite{DBLP:conf/icml/ParkKKD25} uses multi-scale object-relevant patches and coarse-to-fine contrastive decoding. TRiCD~\cite{DBLP:journals/corr/abs-2602-00559} targets video hallucinations through temporal perturbations and saliency-guided visual enhancement. As reported in Table~\ref{tab:hallucination_mitigation}, none of the three 
methods yields substantial improvement. TRiCD achieves the largest gain ($+1.16$ pp, reaching $9.80\%$ avg), with positive deltas across all five question types, suggesting that its video-native contrastive signal is best aligned with the multi-frame temporal structure of corner-case events. While SECOND provides a marginal improvement ($+0.36$ pp avg), VDGD hurts performance ($-1.41$ pp avg), suggesting that description-grounded decoding does not transfer well to dynamic, multi-frame driving inputs. Qualitative examples in Figure~\ref{fig:mitigation} show several flip cases where these decoding methods can either improve or degrade the quality of scene descriptions. Implementation details are in Appendix~\ref{appsec:hallucination_mitigation}.

Though these methods have shown strong results on object-level hallucination with multiple-choice verification, corner-case 
driving evaluation is inherently harder, targeting temporally 
concentrated events and based on open-ended generation, a setting in which 
Video-LLMs are known to exhibit substantially high hallucination rates~\cite{DBLP:conf/iccv/RawalSHSG25}.Taken together, these results on \internvlthreefive indicate that existing training-free hallucination 
mitigation methods provide, at best, marginal and inconsistent benefits on safety-critical corner cases.

\begin{table}[htbp]
\centering
\scriptsize
\renewcommand{\arraystretch}{1.1}
\newcolumntype{C}{>{\centering\arraybackslash}X}
\begin{tabularx}{\linewidth}{l|CCCCC|C|C}
\hline
\textbf{Model} &
\textbf{Q1} & \textbf{Q2} & \textbf{Q3} &
\textbf{Q4\textsubscript{R}} & \textbf{Q5} &
\textbf{Avg Agr.} &
\textbf{Q4\textsubscript{TF}} \\
\hline
\internvlthreefive~\texttt{(8B)} & 4.27 & 7.67 & 6.55 & 12.17 & 12.52 & 8.64 & 78.17 \\
\rowcolor{gray!15}
+\,VDGD~\cite{DBLP:conf/iclr/GhoshEKTNJM25}   & $4.10\pminus{0.17}$ & $5.32\pminus{2.35}$ & $6.96\pplus{0.41}$  & $9.13\pminus{3.04}$  & $10.65\pminus{1.87}$ & $7.23\pminus{1.41}$ & $58.46\pminus{19.71}$ \\
+\,SECOND~\cite{DBLP:conf/icml/ParkKKD25}      & $4.04\pminus{0.23}$ & $7.55\pminus{0.12}$ & $7.14\pplus{0.59}$  & $13.40\pplus{1.23}$  & $12.87\pplus{0.35}$  & $9.00\pplus{0.36}$  & $81.33\pplus{3.16}$ \\
\rowcolor{gray!15}
+\,TRiCD~\cite{DBLP:journals/corr/abs-2602-00559} & $5.03\pplus{0.76}$ & $8.72\pplus{1.05}$ & $7.20\pplus{0.65}$ & $13.34\pplus{1.17}$ & $14.69\pplus{2.17}$ & $9.80\pplus{1.16}$ & $75.13\pminus{3.04}$ \\
\hline
\end{tabularx}
\vspace{6pt}
\caption{Accuracy (\%) of training-free decoding methods applied to
\internvlthreefive~\texttt{(8B)} on the full DashBench. Subscripts show the change in score:
\textcolor{green!55!black}{green} = improvement,
\textcolor{red!80!black}{red} = regression.}
\label{tab:hallucination_mitigation}
\end{table}

\section{Discussion and Conclusion}

Our results reveal that the latest Video LLMs, regardless of open vs. proprietary
status, model scale, reasoning capability, or fine-tuning on established driving datasets, fall critically short when tasked with understanding corner cases. No evaluated model
surpasses 50\% on any individual prompt, and none exceeds 35\% on average:
\gptfive, the highest performer, reaches only 46.52\% on Q4\textsubscript{R} and
34.75\% average agreement. Open-source models are further behind still, with the strongest open-source result (\qwenvl \texttt{235B}, 20.67\% average) lagging the closed-source best by over 14 percentage points. When considering model training recipes neither extended chain-of-thought reasoning nor domain-specific driving specialization on existing datasets narrows this gap, as specialist models fine-tuned on AD datasets achieve at most 10.01\% average agreement, with several performing near chance. More critically, all Video LLMs frequently hallucinate nonexistent road agents, fabricate
imaginary actions, or misclassify collisions as near-misses in addition to simply missing
the evident corner case on the road. These failure modes direct safety risks when Video
LLMs are embedded as explainability or planning components in an ADS.

Our benchmark design and additional experiments conducted further expose the depth of the problem. Prompt formulation is far from a trivial design choice when deploying Video LLMs for road scene analysis, as the expectation that models can robustly describe scenes without being cued to look for anomalies does not hold. We also explored whether narrowing the temporal context to the anomalous event itself could help: isolating the corner-case window does benefit stronger models by several percentage points (+3.79 pp for \gptfive), but the gains for open-source models are negligible (+0.81--1.98 pp), ruling out insufficient visual context as the primary obstacle. We further attempt to address these failures by applying three recent hallucination mitigation methods,with the best approach gaining only $+1.16$ pp on average. Together, these findings indicate that closing the corner-case gap requires fundamentally new approaches: future Video LLM assistants and planners should demonstrate robust visual grounding and must be exposed to diverse scenarios during training. Additionally, motivated by the scarcity of rare non-collision corner cases 
in naturalistic footage, we believe that the development of driving-domain 
world models that can synthesize diverse, controllable 
safety-critical scenarios is a promising direction for future dataset construction.

\textbf{DashBench} provides the most comprehensive ego-view corner-case evaluation of Video LLMs to date: 1,709 dashcam clips across 88 fine-grained categories, balanced across collision and non-collision events, with open-ended free-text QA evaluated by a reliable, SOTA LLM-as-a-judge. Together with a detailed characterization of systematic failure modes and a suite of diagnostic experiments, we offer both a rigorous stress test and a practical guide for the limitations that must be overcome. We believe DashBench serves as a foundational resource for advancing Video LLMs toward the perceptual robustness and safety-critical reliability that trustworthy autonomous driving demands.

\section*{Acknowledgements}
This work was supported by the National Research Foundation of Korea(NRF) grant funded by the Korea government(MSIT)(RS-2025-00573160), and the “Advanced GPU Utilization Support Program” funded by the Government of the Republic of Korea (Ministry of Science and ICT).

\bibliography{bmvc_dashbench}

\clearpage
\appendix

\section*{\LARGE Supplementary Material: Evaluating Video LLMs’ Understanding of
Corner Cases in Autonomous Driving}

\vspace{50pt}
\section{Models and Access Details}
\label{appsec:models}

In this section, we provide the implementation and deployment details for all evaluated Video LLMs used in \textbf{DashBench}.

\subsection{Closed-source Video LLMs}

We evaluate 2 closed-source Video LLMs through their official APIs. 

\subsubsection{\gptfive}
We access \gptfive via the OpenAI API at \url{platform.openai.com} with the following settings:
\begin{itemize}
\item Reasoning effort: \texttt{medium} (default)
  \item Model: \texttt{gpt-5-2025-08-07}
\end{itemize}

\subsubsection{\geminiflash}
We access \geminiflash via OpenRouter at \url{https://openrouter.ai/google/gemini-3-flash-preview} with the following settings:
\begin{itemize}
  \item Thinking mode: \texttt{high} (default)
  \item Model: \geminiflash
\end{itemize}

\subsection{Open-source and Specialist Video LLMs}

We evaluate 23 open-source Video LLMs deployed locally using vLLM (when supported) or HuggingFace Transformers. For all models, we use these default parameters: \texttt{max\_new\_tokens} = 512, \texttt{do\_sample} = False, \texttt{temperature} = 0.0. Exact details related to hardware inference are detailed in the following section.

\subsection{\oThree for AI-Assisted Ground Truth Annotation}

We evaluate \oThree via the OpenAI API at \url{platform.openai.com} with the following settings:
\begin{itemize}
  \item reasoning\_effort: medium (default thinking mode setting)
  \item Model: \texttt{o3-2025-04-16}
\end{itemize}

\subsection{Hardware Configuration}

Experiments for 2 230B+ models (\qwenvl \texttt{235B}, \internvlthreefive \texttt{241B}) and the model as well as methods involved with hallucination mitigation were conducted using the following hardware configuration:

\begin{itemize}
    \item GPU: 4 x Nvidia B200
    \item CPU: Dual-socket Intel Xeon 6960P — 144 cores / 288 threads
    \item RAM: 2 TB
    \item CUDA: 13.0
\end{itemize}

Experiments for all 7 specialist models  were conducted using the following hardware configurations:

\begin{itemize}
    \item GPU: 1 x Nvidia 5090
    \item CPU: AMD Ryzen 9 9900X (12 cores / 24 threads)
    \item RAM: 113 GB
    \item CUDA: 12.8
\end{itemize}

The remaining experiments for all other models are run with the following hardware configuration:

\begin{itemize}
    \item GPU: 4 x Pro 6000
    \item CPU: Dual-socket AMD EPYC 9654 — 112 cores / 224 threads
    \item RAM: 462 GB
    \item CUDA: 12.8
\end{itemize}

\section{Prompts}
\label{appsec:prompts}

In this section, we provide details for the prompts used for different components of our benchmark. The ground truth annotation prompt (Figure~\ref{fig:gt_prompt}) was used with \oThree to assist in the process of creating descriptive annotations for each scenario. The evaluation prompts (Figure~\ref{fig:all_prompt}), detailed in the main paper, were passed to the Video LLMs to test their responses across a range of question types, along with a system prompt (Figure~\ref{fig:system_prompt}). Finally, the judge prompt (Figure~\ref{fig:judge_prompt}) was used to instruct \deepseekvthree to compare model outputs against the ground truth and determine alignment.

\clearpage
\begin{figure*}[t]
\centering
\begin{AIboxFull}{Ground truth annotation prompt}
    \smallskip
    \begin{center}
        \begin{quote}
            \texttt{Look at these 3 frames from a dashcam video. Describe in one sentence what unusual or dangerous event is visible. The unusual or dangerous event happening in this scene is: \{description\}.}
            \end{quote}
    \end{center}
    \smallskip
\end{AIboxFull}
\caption{The prompt passed to \oThree for assisting in ground truth annotation. \emph{\{description\}} is the corner case category.}
\label{fig:gt_prompt}
\end{figure*}

\begin{figure*}[h!]
\centering
\begin{AIboxFull}{Judge Prompt}
    \smallskip
    \begin{center}
        \begin{quote}
        \texttt{Your task is to decide whether the model-generated description captures the same core unusual or dangerous road event as the ground truth.Focus specifically on whether the model matches the core cause, effect, and timing of the danger. Descriptions should agree on what unusual thing happened, when it happened, and why it was dangerous.\\
        If the model omits the suddenness of the action, a major agent involved, or misrepresents who caused the danger, mark it as a mismatch.\\
        You may ignore small differences in phrasing or object types, but do not accept answers that are vague, incomplete, or shift the core focus of the event.\\
        First analyze, then report the final answer in either of Yes or No followed by the detailed explanation.\\
        Examples:\\
        Yes (if the model captures the same unusual or dangerous event\\
        ---detailed explanation---\\
        No (if it misses or distorts the key causual or temporal element)\\
        ---detailed explanation---\\
        \medskip
        Model response: \{model\_response\}\\
        \medskip
        Ground truth: \{ground\_truth\}}
        \end{quote}
    \end{center}
    \smallskip
\end{AIboxFull}
\caption{The prompt passed to \deepseekvthree for judging. \emph{\{model\_response\}} and \emph{\{ground\_truth\}} contains the evaluated Video LLM's response for a specific scenario and the manually created ground truth annotation for the same scenario in the given order.}
\label{fig:judge_prompt}
\end{figure*}

\clearpage

\begin{figure*}[h!]
\centering
\begin{AIboxFull}{Detailed prompts for \textbf{DashBench}}
    \smallskip
    \begin{center}
        \begin{enumerate}[label=\textbf{(Q\arabic*)},  %
                 labelsep=0.6em,               %
                 align=left,                   %
                 leftmargin=*]                 %
        \item \begin{quote}
            \texttt{What is unusual in this sequence of frames?}
            \end{quote}
        \item \begin{quote}
            \texttt{What is dangerous in this sequence of frames?}
            \end{quote}
        \item \begin{quote}
            \texttt{Describe this sequence of frames in detail.}
            \end{quote}
        \item \begin{quote}
            \texttt{You are an expert in analyzing dash‑cam footage and should determine if the sequence of frames depicts an unusual or dangerous event on the road, or if it is a normal road situation. What unusual or dangerous event is happening in this scene?}\\[2pt]
      \texttt{Provide your answer in the following JSON format:}\\
      \texttt{\{}\\
      \texttt{\quad"reasoning": "Brief explanation of why you think this is an unusual or dangerous event or if it is a normal situation.",}\\
      \texttt{\quad"unusual\_or\_dangerous\_event\_detected": true/false}\\
      \texttt{\}}
            \end{quote}
        \item \begin{quote}
            \qfive
        \end{quote}
        \end{enumerate}
        
    \end{center}
    \smallskip
\end{AIboxFull}
\caption{The extensive prompts passed to test the Video LLMs in our benchmark.}
\label{fig:all_prompt}
\end{figure*}

\begin{figure*}[h!]
\centering
\begin{AIboxFull}{System prompt}
    \smallskip
    \begin{center}
        \begin{quote}
            \texttt{You are a helpful driving-scene assistant. Please be brief and concise in your description. Answer within 250 words.}
            \end{quote}
    \end{center}
    \smallskip
\end{AIboxFull}
\caption{The system prompt passed to the Video LLM along with the evaluation prompts.}
\label{fig:system_prompt}
\end{figure*}

\begin{table*}[p]
  \centering
  \footnotesize
  \renewcommand{\arraystretch}{1.1}
  \begin{tabularx}{\textwidth}{|X|X|}
    \hline
    \textbf{Collision Categories (46 types, 890 samples)} & \textbf{Non-Collision Categories (42 types, 819 samples)} \\
    \hline
    \small
    car hits a curb (32) & car has failure of components (22) \\
    car hits car (30) & car is out of control (105) \\
    car hits cyclist (32) & cyclist falls down (6) \\
    car hits motorbike (30) & ego-car has failure of components (1) \\
    car hits pedestrian (31) & ego-car is out of control (81) \\
    car hits telegraphpoles (22) & ego-car is overtaken by a car (105) \\
    car hits trees (23) & ego-car is overtaken by a truck (29) \\
    car scratches truck (28) & motorbike falls down (45) \\
    cyclist hits cyclist (2) & pedestrian falls down (3) \\
    ego-car hits a car (28) & truck has failure of components (42) \\
    ego-car hits a crossing car (29) & truck is out of control (66) \\
    ego-car hits a crossing cyclist (18) & \underline{dangerous lane change/merging (44)} \\
    ego-car hits a crossing motorbike (31) & \underline{car drives against traffic flow (36)} \\
    ego-car hits a crossing pedestrian (29) & \underline{car runs a red light without collision (31)} \\
    ego-car hits a crossing truck (25) & \underline{pedestrian hazard (23)} \\
    ego-car hits a curb (29) & \underline{dangerous overtaking (17)} \\
    ego-car hits a cyclist (11) & \underline{motorbike drives against traffic flow (16)} \\
    ego-car hits a motorbike (29) & \underline{improper stopping on the road (15)} \\
    ego-car hits a pedestrian (7) & \underline{animal hazard (12)} \\
    ego-car hits a truck (27) & \underline{near-miss (11)} \\
    ego-car hits telegraphpoles (9) & \underline{parking/roadside stopping hazard (11)} \\
    ego-car hits trees (22) & \underline{slippery road conditions due to water on the road (10)} \\
    motorbike hits a curb (2) & \underline{large object dropped from vehicle (9)} \\
    motorbike hits motorbike (22) & \underline{vehicle collapses onto road (9)} \\
    motorbike hits pedestrian (3) & \underline{bus drives against traffic flow (8)} \\
    motorbike scratches car (24) & \underline{truck drives against traffic flow (7)} \\
    motorbike scratches truck (1) & \underline{improper turning (6)} \\
    truck hits a curb (29) & \underline{dangerous car reversing (5)} \\
    truck hits car (28) & \underline{motorbike runs a red light without collision (5)} \\
    truck hits cyclist (3) & \underline{slippery road conditions due to snow (5)} \\
    truck hits motorbike (31) & \underline{vehicle goes off-road (5)} \\
    truck hits pedestrian (4) & \underline{road obstruction \& blockage (4)} \\
    truck hits telegraphpoles (4) & \underline{cyclist violating traffic rules (3)} \\
    truck hits trees (2) & \underline{emergency vehicle interaction (3)} \\
    truck hits truck (30) & \underline{flooded road segment (3)} \\
    car hits large roadblocks (31) & \underline{limited visibility from structure/object (3)} \\
    car hits other road facilities (31) & \underline{road surface hazard (3)} \\
    car hits small roadblocks/road pothholes (15) & \underline{burning object hazard (2)} \\
    ego-car hits large roadblocks (30) & \underline{crime/safety threats on the road (2)} \\
    ego-car hits other road facilities (12) & \underline{structure collapses onto road (2)} \\
    ego-car hits small roadblocks/road pothholes (27) & \underline{tree collapses onto road (2)} \\
    motorbike hits large roadblocks (2) & \underline{truck runs a red light without collision (2)} \\
    motorbike hits other road facilities (1) & \\
    truck hits large roadblocks (30) & \\
    truck hits other road facilities (2) & \\
    truck hits small roadblocks/road pothholes (2) & \\
    \hline
  \end{tabularx}
  \vspace{9pt}
  \caption{Categorization of corner cases in \textbf{DashBench} (1,709 total samples, 88 categories). Collision categories are on the left (46 types, 890 samples) and non-collision categories on the right (42 types, 819 samples). \underline{Underlined} categories are sourced from the RoadSocial dataset.}
  \label{tab:categorization}
\end{table*}

\clearpage

\begin{table*}[!t]
  \centering
  \footnotesize
  \setlength{\tabcolsep}{3pt}
  \renewcommand{\arraystretch}{1.1}
  \begin{tabularx}{\textwidth}{|Xr|Xr|}
    \hline
    \multicolumn{2}{|l|}{\textbf{Collision Categories (46 types, 890 samples)}} &
    \multicolumn{2}{l|}{\textbf{Non-Collision Categories (42 types, 819 samples)}} \\
    \hline

    car hits a curb & \textbf{32.50\%} &
    car has failure of components & \textbf{20.00\%} \\

    car hits car & \textbf{9.33\%} &
    car is out of control & \textbf{31.05\%} \\

    car hits cyclist & \textbf{31.25\%} &
    cyclist falls down & \textbf{23.33\%} \\

    car hits motorbike & \textbf{28.00\%} &
    ego-car has failure of components & \textbf{0.00\%} \\

    car hits pedestrian & \textbf{21.29\%} &
    ego-car is out of control & \textbf{22.47\%} \\

    car hits telegraphpoles & \textbf{26.36\%} &
    ego-car is overtaken by a car & \textbf{11.24\%} \\

    car hits trees & \textbf{18.26\%} &
    ego-car is overtaken by a truck & \textbf{9.66\%} \\

    car scratches truck & \textbf{6.43\%} &
    motorbike falls down & \textbf{29.33\%} \\

    cyclist hits cyclist & \textbf{20.00\%} &
    pedestrian falls down & \textbf{46.67\%} \\

    ego-car hits a car & \textbf{5.71\%} &
    truck has failure of components & \textbf{21.43\%} \\

    ego-car hits a crossing car & \textbf{11.03\%} &
    truck is out of control & \textbf{39.09\%} \\

    ego-car hits a crossing cyclist & \textbf{16.67\%} &
    \underline{dangerous lane change/merging} & \textbf{16.36\%} \\

    ego-car hits a crossing motorbike & \textbf{11.61\%} &
    \underline{car drives against traffic flow} & \textbf{1.67\%} \\

    ego-car hits a crossing pedestrian & \textbf{22.07\%} &
    \underline{car runs a red light without collision} & \textbf{18.71\%} \\

    ego-car hits a crossing truck & \textbf{9.60\%} &
    \underline{pedestrian hazard} & \textbf{29.57\%} \\

    ego-car hits a curb & \textbf{3.45\%} &
    \underline{dangerous overtaking} & \textbf{0.00\%} \\

    ego-car hits a cyclist & \textbf{1.82\%} &
    \underline{motorbike drives against traffic flow} & \textbf{2.50\%} \\

    ego-car hits a motorbike & \textbf{3.45\%} &
    \underline{improper stopping on the road} & \textbf{8.00\%} \\

    ego-car hits a pedestrian & \textbf{11.43\%} &
    \underline{animal hazard} & \textbf{63.33\%} \\

    ego-car hits a truck & \textbf{6.67\%} &
    \underline{near-miss} & \textbf{7.27\%} \\

    ego-car hits telegraphpoles & \textbf{4.44\%} &
    \underline{parking/roadside stopping hazard} & \textbf{7.27\%} \\

    ego-car hits trees & \textbf{24.55\%} &
    \underline{slippery road conditions due to water on road} & \textbf{26.00\%} \\

    motorbike hits a curb & \textbf{0.00\%} &
    \underline{large object dropped from vehicle} & \textbf{20.00\%} \\

    motorbike hits motorbike & \textbf{10.00\%} &
    \underline{vehicle collapses onto road} & \textbf{37.78\%} \\

    motorbike hits pedestrian & \textbf{0.00\%} &
    \underline{bus drives against traffic flow} & \textbf{5.00\%} \\

    motorbike scratches car & \textbf{8.33\%} &
    \underline{truck drives against traffic flow} & \textbf{2.86\%} \\

    motorbike scratches truck & \textbf{0.00\%} &
    \underline{improper turning} & \textbf{23.33\%} \\

    truck hits a curb & \textbf{30.34\%} &
    \underline{dangerous car reversing} & \textbf{0.00\%} \\

    truck hits car & \textbf{13.57\%} &
    \underline{motorbike runs a red light without collision} & \textbf{20.00\%} \\

    truck hits cyclist & \textbf{6.67\%} &
    \underline{slippery road conditions due to snow} & \textbf{8.00\%} \\

    truck hits motorbike & \textbf{15.48\%} &
    \underline{vehicle goes off-road} & \textbf{56.00\%} \\

    truck hits pedestrian & \textbf{5.00\%} &
    \underline{road obstruction \& blockage} & \textbf{30.00\%} \\

    truck hits telegraphpoles & \textbf{10.00\%} &
    \underline{cyclist violating traffic rules} & \textbf{0.00\%} \\

    truck hits trees & \textbf{0.00\%} &
    \underline{emergency vehicle interaction} & \textbf{33.33\%} \\

    truck hits truck & \textbf{9.33\%} &
    \underline{flooded road segment} & \textbf{86.67\%} \\

    car hits large roadblocks & \textbf{32.90\%} &
    \underline{limited visibility from structure/object} & \textbf{6.67\%} \\

    car hits other road facilities & \textbf{28.39\%} &
    \underline{road surface hazard} & \textbf{6.67\%} \\

    car hits small roadblocks/road pothholes & \textbf{8.00\%} &
    \underline{burning object hazard} & \textbf{50.00\%} \\

    ego-car hits large roadblocks & \textbf{9.33\%} &
    \underline{crime/safety threats on the road} & \textbf{40.00\%} \\

    ego-car hits other road facilities & \textbf{10.00\%} &
    \underline{structure collapses onto road} & \textbf{0.00\%} \\

    ego-car hits small roadblocks/road pothholes & \textbf{6.67\%} &
    \underline{tree collapses onto road} & \textbf{70.00\%} \\

    motorbike hits large roadblocks & \textbf{0.00\%} &
    \underline{truck runs a red light without collision} & \textbf{10.00\%} \\

    motorbike hits other road facilities & \textbf{0.00\%} &
    & \\

    truck hits large roadblocks & \textbf{24.67\%} &
    & \\

    truck hits other road facilities & \textbf{50.00\%} &
    & \\

    truck hits small roadblocks/road pothholes & \textbf{30.00\%} &
    & \\

    \hline
  \end{tabularx}
  \vspace{9pt}
  \caption{
    Category-wise accuracy (\%) of \qwenvl{} 8B on the DashBench corner-case categories.
    Accuracy is computed by pooling the Q1--Q5 results within each category.
    Collision categories are shown on the left (46 types, 890 samples), while
    non-collision categories are shown on the right (42 types, 819 samples).
    \underline{Underlined} categories are sourced from the RoadSocial dataset.
  }
  \label{tab:category_accuracy}
\end{table*}

\clearpage

\begin{figure}[h!]
\centering
\includegraphics[width=0.5\textwidth]{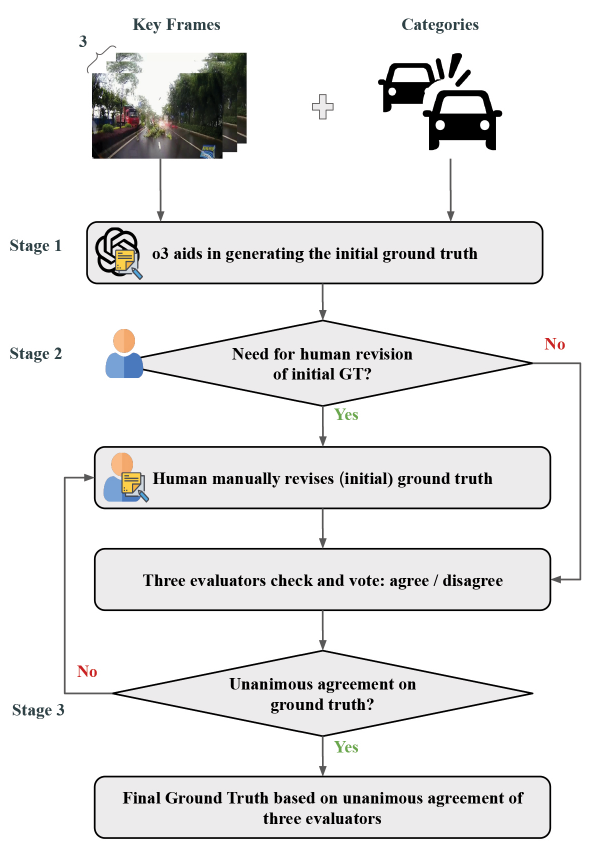}
\vspace{9pt}
\caption{Illustration of our complete pipeline for o3-assisted scene annotation as well as manual annotation review and validation for reliable ground truth descriptions.}
\label{fig:valid_pipeline}
\end{figure}

\section{Ground Truth Annotation}
\label{app:gt_annotation}

\subsection{Assisted Scene Annotation with \texttt{\oThree}}

Since writing detailed scene descriptions for hundreds of clips is labour-intensive, we leverage OpenAI's \texttt{\oThree} model to generate detailed scene-level annotations. The process for generating initial ground truth annotations using \oThree's assistance for real corner case scenarios involve these steps:

\textbf{Keyframe extraction}
For each clip, we extract three critical keyframes representing the progression of the anomalous event. These frames correspond to: (1) the start of the video, (2) the corner case occurrence frame (when the scene reaches its most critical point), and (3) end of the video. This temporal breakdown allows the model to observe the evolution of each incident, though with a limited context.

\textbf{Contextual prompting}
Along with the keyframes, we provide the corner-case category label as a concise answer to the model. This helps the model ground its scene-level annotation based on the given category answer, encouraging it to focus on relevant visual semantics and avoid generic captions.

\textbf{Scene description generation}
Using these inputs, the \texttt{\oThree} model generates descriptive captions as prompted. The detailed prompt can be seen in Figure \ref{fig:gt_prompt}. These captions capture both object-level details and event-level dynamics. 

\subsection{Manual Annotation Review and Validation}

However, even with passed keyframes and the descriptive hint from the corner case category for creating the ground truth annotation, \oThree might fail to create a descriptive annotation based on what actually happened in the scene since its context is limited and it might misinterpret key details. To ensure accuracy and reliability of automatically generated annotations, we implement a multi-phase validation pipeline involving human annotators. This process verifies that all labels and descriptions reflect the visual evidence, align with intended semantics, and maintain consistency across the dataset.

\begin{figure}[h!]
\centering
\includegraphics[width=0.8\linewidth]{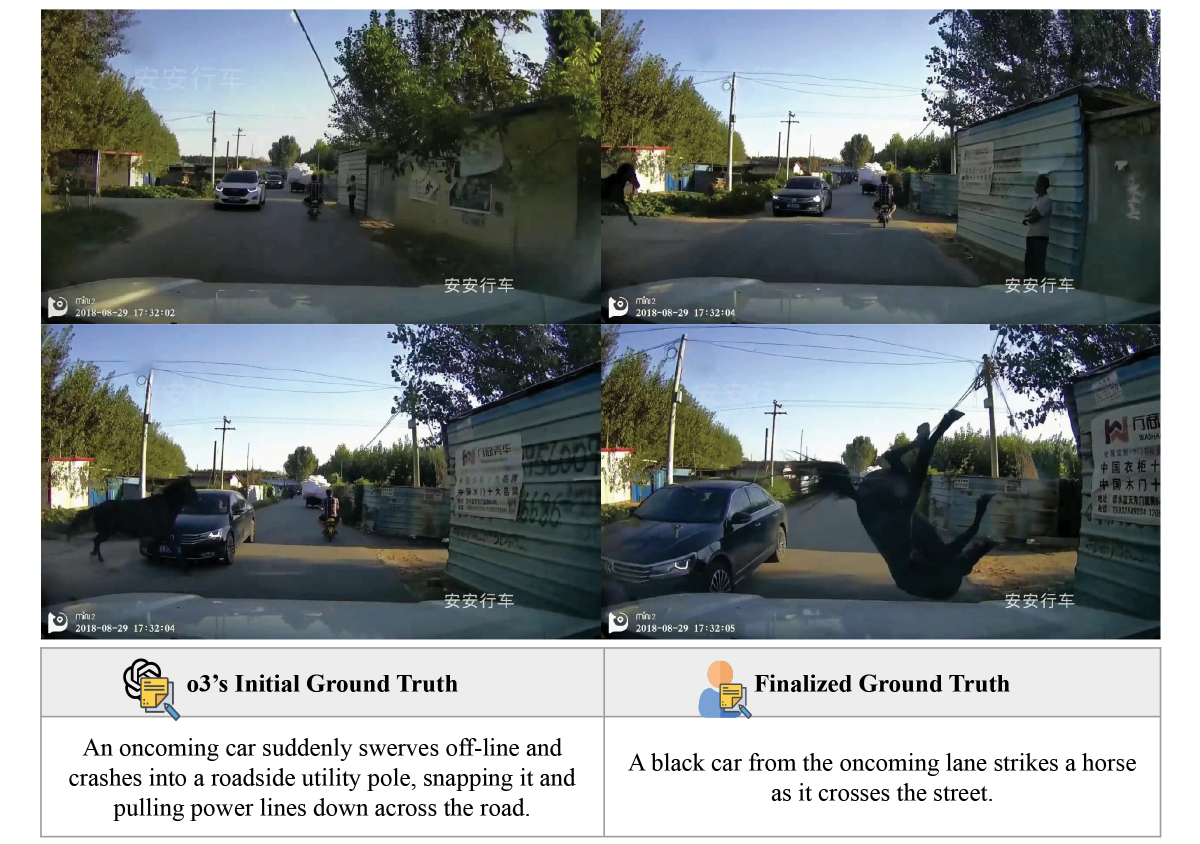}
\vspace{9pt}
\caption{An example corner case scenario where o3's initial scene annotation is inaccurate and misses key details. Through our pipeline, we ensure that all samples in our dataset go through extensive manual revision where necessary to ensure annotation quality.}
\label{fig:validation_ex}
\end{figure}

\textbf{Validation pipeline} As shown in Figure \ref{fig:valid_pipeline}, our annotation refinement pipeline follows a systematic approach:
\begin{enumerate}
    \item \textbf{Initial manual review} All scene descriptions generated by \oThree are first reviewed by a human annotator. The annotator checks for factual errors, hallucinations, inconsistencies, or ambiguities in the captions, by comparing them with the \textbf{full list of frames for the specific sample.} With the full context, minor phrasing issues or missing details are corrected at this stage to enhance clarity and precision.

    \item \textbf{Evaluator panel verification} After initial manual revision, a panel of three evaluators examines the refined annotations and also compares with the full list of frames for the specific sample. Each evaluator independently assesses whether the caption accurately and adequately describes the visual content and evolving corner case situation in the specific scenario.

    \item \textbf{Unanimous agreement check} We adopt a strict unanimity principle for final ground-truth acceptance. The revised ground truth annotation is approved only if all three evaluators independently agree on its validity. This high consensus bar ensures robustness and minimizes individual bias or interpretation errors.

    \item \textbf{Conflict resolution} If full agreement is not reached, the sample is flagged as "in conflict." These cases are brought back to the initial human annotator for additional human revision and will pass through the evaluator panel again.

    \item \textbf{Final ground truth consolidation} Once consensus is achieved through unanimous agreement, the validated ground truth annotation is finalized. These annotations form the basis of the benchmark's evaluation targets, ensuring trustworthiness and reproducibility for future research. An example corner case scenario with the initial \oThree-generated initial annotation and the revised, final ground-truth annotation is shown in Figure \ref{fig:validation_ex}.
\end{enumerate}

\section{LLM-as-a-Judge Evaluation}
\label{appsec:evaluation}

Reliable benchmarking needs an automatic judge whose decisions mirror human judgments. Following existing works using the LLM-as-judge setup, we try out several candidate LLM-as-judges for our work, namely \deepseekvthree, \llama, and also a setup using multiple LLM-as-judges as an ensemble \cite{DBLP:journals/corr/abs-2404-18796}, named PoLL. Specifically, PoLL uses three relatively cost-efficient LLM-as-judges: \texttt{Command-R}, \texttt{Claude 3 Haiku}, and \texttt{GPT-3.5 Turbo}. This ensemble setup is designed to reduce intra-model bias as PoLL aggregates panel votes via \emph{max voting}: the panel labels a response as correct if \emph{at least one} judge deems it correct. For each candidate judge, we randomly sample 100 clips and run evaluation on five different models as shown in Table \ref{tab:judge_kappa}, obtain the judge's decision, and compare it 
against human evaluation of whether the Video LLM's response is aligned with the ground truth. To ensure the human reference is itself reliable, we follow a modified consensus-based validation pipeline:

\begin{enumerate}
  \item \textbf{Candidate Judge Generates Initial Judgment.} For each sample and Video LLM output, 
  the candidate judge independently labels the response as correct or incorrect, based solely 
  on the Video LLM's response and the ground-truth annotation. The prompt used is shown in Figure~\ref{fig:judge_prompt}.

  \item \textbf{Evaluator Panel Alignment Check} A panel of three human evaluators reviews each sample, considering the same textual output from the Video LLM and the ground truth. Each evaluator independently decides whether the Video LLM’s answer is aligned with the ground truth.

  \item \textbf{Unanimous Agreement Check} We require \emph{full consensus}, meaning that all three evaluators must agree among themselves on the correct/incorrect label for whether or not the model response is aligned with the ground truth annotation. In all other cases, those samples are discussed among all three evaluators to reach a final decision. Only then do we compare this unanimous human consensus with the candidate judge's label.

  \item \textbf{Final Agreement and Selection.} Cohen's $\kappa$ is computed between each candidate 
  judge and the unanimous human consensus. As shown in Table \ref{tab:judge_kappa}, \deepseekvthree achieves the highest agreement 
  ($\kappa = 0.96$), confirming it closely mirrors human judgement and can serve as a dependable LLM-as-judge for our benchmark. 
  
\end{enumerate}

\subsection{Judge Access Details}

For each LLM-as-judge mentioned, we access the model via OpenRouter at \url{https://openrouter.ai/} and run judging with the judge prompt shown in Figure \ref{fig:judge_prompt} with temperature = 0.0.

\subsection{Judge Variance Across Runs}

\begin{table}[!htbp]
\centering
\footnotesize
\setlength{\tabcolsep}{3pt}
\renewcommand{\arraystretch}{1.1}
\begin{tabular}{l|ccccc|c}
\hline
\multirow{2}{*}{\textbf{Model}} &
\multirow{2}{*}{\centering\textbf{Q1}} &
\multirow{2}{*}{\centering\textbf{Q2}} &
\multirow{2}{*}{\centering\textbf{Q3}} &
\multirow{2}{*}{\centering\textbf{Q4\textsubscript{R}}} &
\multirow{2}{*}{\centering\textbf{Q5}} &
\multirow{2}{*}{\parbox{1.2cm}{\centering\textbf{Avg}\\\textbf{Agree}}} \\
& & & & & & \\ \hline
\multicolumn{7}{l}{\textbf{Closed-source}} \\
\hline
\gptfive{}       & $20.32_{\pm 0.27}$ & $36.75_{\pm 0.33}$ & $30.91_{\pm 0.30}$ & $46.53_{\pm 0.30}$ & $39.82_{\pm 0.67}$ & $\mathbf{34.86_{\pm 0.24}}$ \\
\geminiflash{}   & $31.50_{\pm 0.61}$ & $34.82_{\pm 0.44}$ & $29.03_{\pm 0.52}$ & $39.16_{\pm 0.55}$ & $34.38_{\pm 0.39}$ & $\mathbf{33.78_{\pm 0.39}}$ \\
\hline
\multicolumn{7}{l}{\textbf{Open-source}} \\
\hline
\qwenvl{} \texttt{(8B)}            & $18.77_{\pm 0.23}$ & $14.92_{\pm 0.38}$ & $13.98_{\pm 0.46}$ & $22.12_{\pm 0.21}$ & $21.52_{\pm 0.15}$ & $\mathbf{18.26_{\pm 0.19}}$ \\
\ovistwofive{} \texttt{(9B)}       & $18.66_{\pm 0.59}$ & $14.57_{\pm 0.57}$ & $\phantom{0}7.99_{\pm 0.22}$ & $17.72_{\pm 0.60}$ & $19.71_{\pm 0.70}$ & $\mathbf{15.64_{\pm 0.39}}$ \\
\internvlthreefive{} \texttt{(8B)} & $\phantom{0}4.29_{\pm 0.15}$ & $\phantom{0}9.49_{\pm 0.39}$ & $\phantom{0}8.39_{\pm 0.32}$ & $16.13_{\pm 0.24}$ & $15.92_{\pm 0.31}$ & $\mathbf{10.84_{\pm 0.23}}$ \\
\hline
\end{tabular}
\vspace{9pt}
\caption{
    \textbf{Judge variance across five independent judging runs.}
     Mean accuracy (\%) $\pm$ sample standard deviation (std) over five runs on identical inputs are reported.  All std $\leq\!0.70$ \%, and model rankings are stable across every individual run.
}
\label{tab:judge_variance}
\end{table}

To measure evaluation stability, we repeat the \deepseekvthree-as-a-judge evaluation pipeline 
across five independent runs on identical inputs. Table~\ref{tab:judge_variance} reports the 
mean accuracy (\%) and sample standard deviation across all five runs, broken down by question 
type (Q1--Q5) and model.

Variance is consistently low across all evaluated models and question types, with standard 
deviations never exceeding 0.70\%. Across the five runs, model rankings remain entirely stable: 
closed-source models (\gptfive and \geminiflash) consistently outperform all open-source 
counterparts, and the relative ordering within each group is preserved in every individual run. These results confirm that \deepseekvthree produces 
deterministic and reproducible judgments at the scale of our benchmark, and that reported 
accuracy figures are not byproducts of stochastic sampling during inference.

\section{Hallucination Mitigation Method Implementation}
\label{appsec:hallucination_mitigation}

All methods are implemented on top of \internvlthreefive~\texttt{(8B)} as the backbone, with 16 uniformly sampled frames resized to $448 \times 448$ via BICUBIC interpolation and normalized with ImageNet mean and standard deviation. Inference uses greedy decoding (\texttt{do\_sample=False}) across all methods, and all prompts are paired with the system prompt shared in Figure \ref{fig:system_prompt}. Common settings are summarized in Table~\ref{tab:hallucination_common}.

\begin{table}[h]
\centering
\footnotesize
\setlength{\tabcolsep}{6pt}
\renewcommand{\arraystretch}{1.1}
\begin{tabular}{ll}
\hline
\textbf{Setting} & \textbf{Value} \\
\hline
Backbone model         & \texttt{OpenGVLab/InternVL3\_5-8B} \\
Conversation template  & \texttt{internvl2\_5} \\
Input frames per video & 16 \\
Frame resolution       & $448 \times 448$ (BICUBIC) \\
Image normalization    & ImageNet mean / std \\
Data type              & \texttt{float16} \\
Decoding strategy      & Greedy (\texttt{do\_sample=False}) \\
GPU parallelism        & 4 (CUDA 0--3, 1:1 shard-to-GPU) \\
Prompt types evaluated & Q1, Q2, Q3, Q4, Q5 \\
\hline
\end{tabular}
\vspace{9pt}
\caption{Common inference settings shared across all hallucination mitigation methods.}
\label{tab:hallucination_common}
\end{table}

\subsection{Baseline}

The baseline runs standard \internvlthreefive~\texttt{(8B)} inference with no auxiliary branches or additional passes. The 16 frames are composed into a multi-image query in the format \texttt{Frame-1: <image> \ldots Frame-16: <image>} and fed through a single greedy forward pass.

\begin{itemize}
    \item \texttt{max\_new\_tokens}: 512
    \item Contrastive branch: none
    \item Additional passes: none
\end{itemize}

\subsection{VDGD (Visual Description-Grounded Decoding)}

VDGD~\cite{DBLP:conf/iclr/GhoshEKTNJM25} uses description-prefixing, operating in two sequential passes over the same 16 frames. The first pass generates a free-text visual description of the clip, and the second pass prepends that description as explicit textual context before the actual question, guiding the model to ground its answer in the described visual content rather than relying on language priors. No contrastive branch is required.

\begin{itemize}
    \item \texttt{max\_new\_tokens} (pass 2, answer): 512
    \item \texttt{desc\_max\_new\_tokens} (pass 1, description): 512
    \item Pass 1 prompt: \texttt{"Describe this sequence of frames in detail."}
    \item Pass 2 prompt format: \texttt{"The video is described as: \{description\}\allowbreak$\backslash$nNow answer the following question.$\backslash$n\{original\_question\}"}
    \item Number of passes: 2
\end{itemize}

\subsection{SECOND}

SECOND~\cite{DBLP:conf/icml/ParkKKD25} implements degraded-context contrast idea where the positive branch receives all 16 frames while the negative branch receives only the single middle frame, intentionally stripping temporal context. The same per-step contrastive formula $\ell_\text{final} = \ell_\text{full} + \alpha \cdot (\ell_\text{full} - \ell_\text{partial})$ is applied to upweight information that is only accessible from the full temporal sequence, suppressing responses that could be generated from a single static frame alone.

\begin{itemize}
    \item \texttt{max\_new\_tokens}: 512
    \item \texttt{alpha} (contrastive weight): 0.5
    \item Positive branch: all 16 original frames
    \item Negative branch: single middle frame (\texttt{paths[len(paths)//2]})
    \item Number of branches: 2 (full / partial)
\end{itemize}

\subsection{TriCD}

TRiCD~\cite{DBLP:journals/corr/abs-2602-00559} employs two branches in parallel: a \emph{positive} branch receiving the original frame sequence and a \emph{negative} branch receiving a randomly shuffled copy (seed\,=\,0), which destroys temporal ordering. At each decoding step, the next-token logits from both branches are combined as $\ell_\text{final} = \ell_\text{pos} + \alpha \cdot (\ell_\text{pos} - \ell_\text{neg})$, encouraging the model to rely on temporal context rather than frame-level statistics. Both branches maintain independent KV-caches synchronized at each step with the same selected token.

\begin{itemize}
    \item \texttt{max\_new\_tokens}: 512
    \item \texttt{alpha} (contrastive weight): 0.5
    \item \texttt{neg\_perturb}: \texttt{shuffle} (fixed seed\,=\,0; alternatives: \texttt{blur}, \texttt{grayscale}, \texttt{noise}, \texttt{reverse})
    \item Number of branches: 2 (positive / negative)
\end{itemize}

\begin{table}[!t]
\centering
\footnotesize
\setlength{\tabcolsep}{3pt}
\renewcommand{\arraystretch}{1.1}
\begin{tabular}{l|ccccc|c|c}
\hline
\multirow{2}{*}{\textbf{Model}} &
\multirow{2}{*}{\centering\textbf{Q1}} &
\multirow{2}{*}{\centering\textbf{Q2}} &
\multirow{2}{*}{\centering\textbf{Q3}} &
\multirow{2}{*}{\centering\textbf{Q4\textsubscript{R}}} &
\multirow{2}{*}{\centering\textbf{Q5}} &
\multirow{2}{*}{\parbox{1.2cm}{\centering\textbf{Avg}\\\textbf{Agree}}} &
\multirow{2}{*}{\centering\textbf{Q4\textsubscript{TF}}} \\
& & & & & & & \\ \hline

\multicolumn{8}{l}{\textbf{Closed-source}} \\
\hline

\gptfive{}
  & 20.90 & 36.09 & 29.28 & 47.39 & 38.40 & 34.41 & 96.96 \\

\rowcolor{gray!15}
\,+\,\textit{CCW}\,16 frames
  & $19.68_{\pminus{1.22}}$
  & $42.41_{\pplus{6.32}}$
  & $34.51_{\pplus{5.23}}$
  & $51.40_{\pplus{4.01}}$
  & $43.01_{\pplus{4.61}}$
  & $38.20_{\pplus{3.79}}$
  & $97.81_{\pplus{0.85}}$ \\

\rowcolor{gray!15}
\,+\,\textit{CCW}\,32 frames
  & $19.20_{\pminus{1.70}}$
  & $40.22_{\pplus{4.13}}$
  & $32.44_{\pplus{3.16}}$
  & $50.43_{\pplus{3.04}}$
  & $43.01_{\pplus{4.61}}$
  & $37.06_{\pplus{2.65}}$
  & $98.18_{\pplus{1.22}}$ \\

\hline
\multicolumn{8}{l}{\textbf{Open-source}} \\
\hline

\qwenvl \texttt{(8B)}
  & 17.98 & 14.95 & 13.97 & 20.90 & 21.02 & 17.76 & 70.84 \\

\rowcolor{gray!15}
\,+\,\textit{CCW}\,16 frames
  & $13.61_{\pminus{4.37}}$
  & $18.35_{\pplus{3.40}}$
  & $15.19_{\pplus{1.22}}$
  & $22.11_{\pplus{1.21}}$
  & $23.57_{\pplus{2.55}}$
  & $18.57_{\pplus{0.81}}$
  & $73.39_{\pplus{2.55}}$ \\

\rowcolor{gray!15}
\,+\,\textit{CCW}\,32 frames
  & $14.58_{\pminus{3.40}}$
  & $20.66_{\pplus{5.71}}$
  & $13.85_{\pminus{0.12}}$
  & $23.45_{\pplus{2.55}}$
  & $22.72_{\pplus{1.70}}$
  & $19.05_{\pplus{1.29}}$
  & $69.50_{\pminus{1.34}}$ \\

\internvlthreefive \texttt{(8B)}
  & \phantom{0}3.65
  & \phantom{0}8.02
  & \phantom{0}6.68
  & 10.57
  & 12.15
  & \phantom{0}8.21
  & 74.48 \\

\rowcolor{gray!15}
\,+\,\textit{CCW}\,16 frames
  & $\phantom{0}4.74_{\pplus{1.09}}$
  & $\phantom{0}9.11_{\pplus{1.09}}$
  & $\phantom{0}7.65_{\pplus{0.97}}$
  & $14.22_{\pplus{3.65}}$
  & $15.19_{\pplus{3.04}}$
  & $10.18_{\pplus{1.97}}$
  & $85.05_{\pplus{10.57}}$ \\

\rowcolor{gray!15}
\,+\,\textit{CCW}\,32 frames
  & $\phantom{0}5.47_{\pplus{1.82}}$
  & $12.88_{\pplus{4.86}}$
  & $\phantom{0}9.84_{\pplus{3.16}}$
  & $15.31_{\pplus{4.74}}$
  & $13.00_{\pplus{0.85}}$
  & $11.30_{\pplus{3.09}}$
  & $80.60_{\pplus{6.12}}$ \\

\hline
\end{tabular}
\vspace{9pt}
\caption{
    Accuracy (\%) of selected models on the CCW subset.
    Each model's full-video run is compared against its CCW runs
    using 16/32 frames sampled from the corner case window only.
    Subscripts show the change in score relative to the corresponding
    full-video run: improvement and regression are indicated by the
    respective signs.
}
\label{tab:ccw_results_expanded}
\end{table}

\section{Additional Results}
\label{appsec:add_results}
In this section, we provide supplementary information and results for the experiments presented in the main paper. First, using \qwenvl{} 8B as a reference, we report category-wise accuracy in Table \ref{tab:category_accuracy}. The grouped results show an accuracy of 15.89\% for collision categories and 21.29\% for non-collision categories, while the MMAU and RoadSocial subsets achieve 18.78\% and 17.13\%, respectively. Additionally, in Table \ref{tab:ccw_results_expanded} we show the results of the corner-case window experiments when not only 16, but also 32 frames, are sampled uniformly only from the corner-case window.

\section{Additional Qualitative Examples}
\label{appsec:qualitative}

\subsection{Failing to Identify Evident Corner Cases}
\label{app:fail_ex}
\begin{figure*}[h!]
\centering
\begin{AIboxFull}{\crashchat misses the motorcycle right in front in the road}
  \begin{center}
  \includegraphics[width=0.50\textwidth]{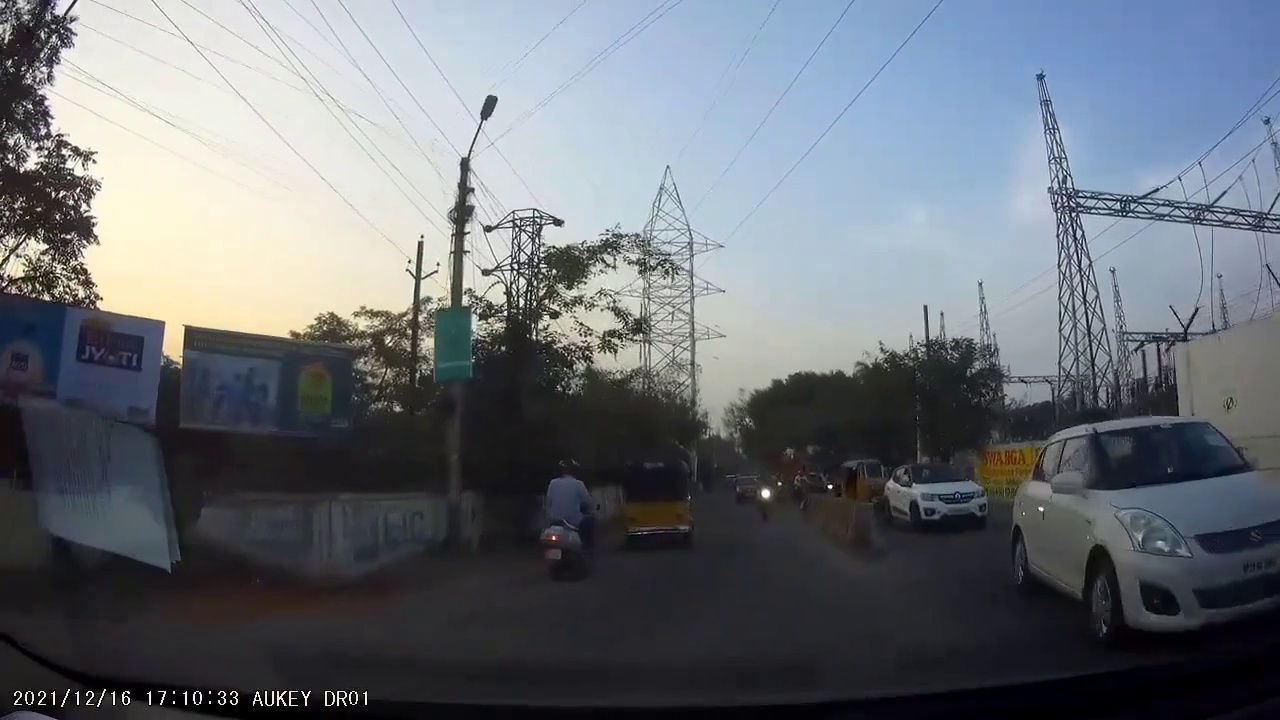}%
  \hfill
  \includegraphics[width=0.50\textwidth]{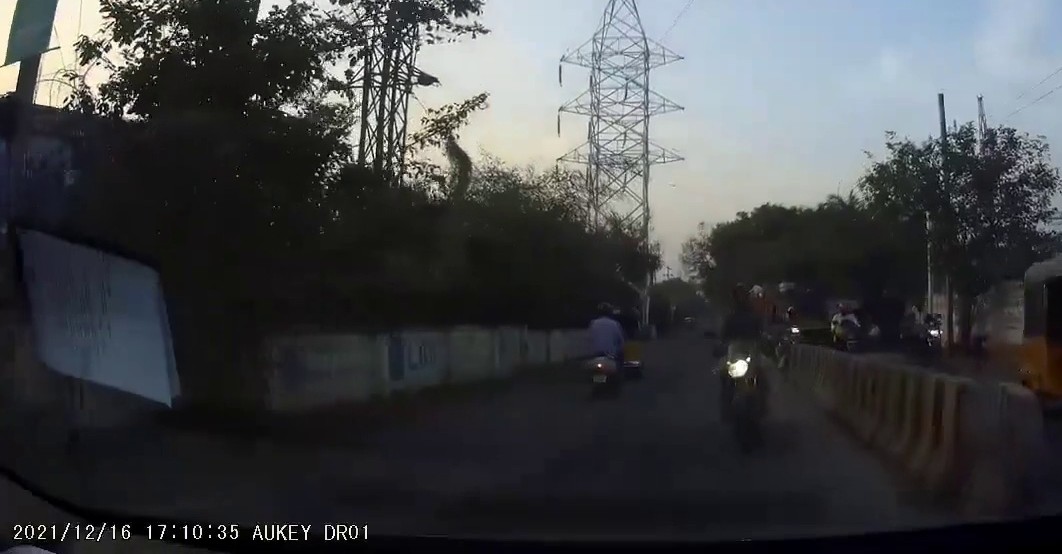}%
  \hfill
  \vspace{-1pt}
  \includegraphics[width=0.50\textwidth]{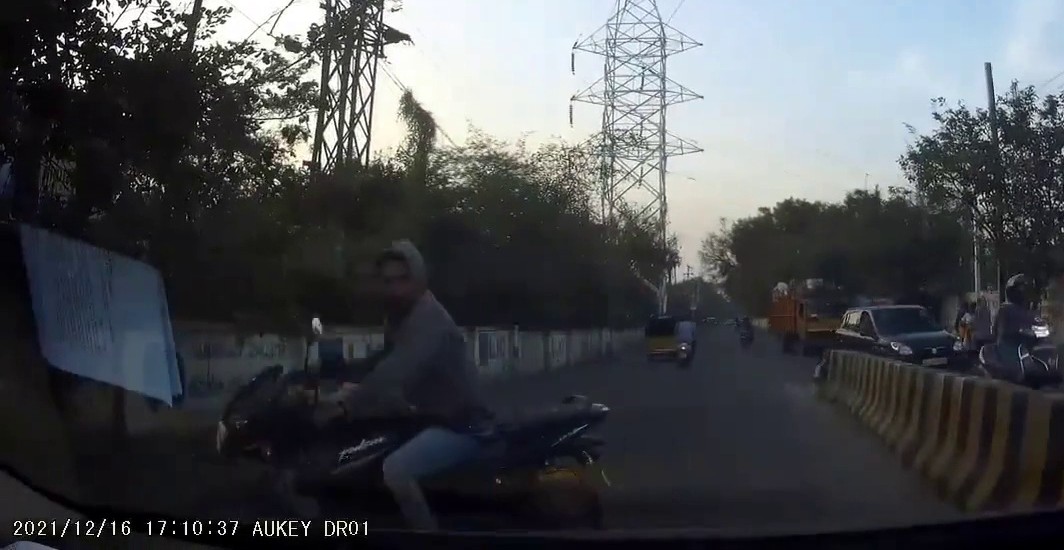}%
  \hfill
  \includegraphics[width=0.50\textwidth]
  {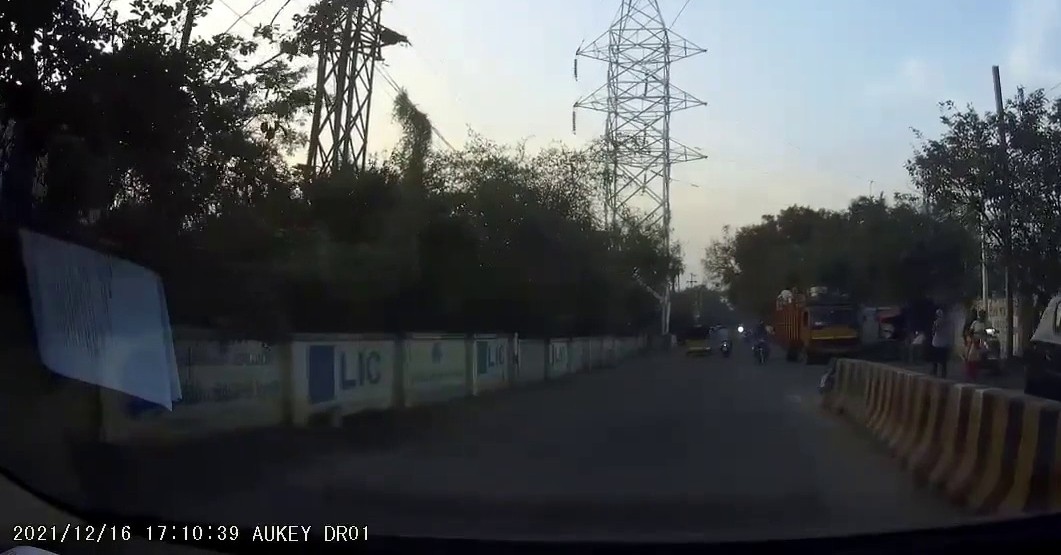}%
  \end{center}
  \medskip

  \textbf{Question Type:} \qthree \\
  \textbf{Ground Truth:} A motorbike is riding directly toward the dash-cam vehicle in the wrong direction of traffic.\\
  \textbf{\crashchat:} The video shows a busy street with various vehicles and people. The camera is mounted on a vehicle, possibly a motorcycle, and is moving forward. In the foreground, there is a yellow and black auto-rickshaw, followed by a white car. On the right side of the road, there are several people walking, including a man in a blue shirt and a woman in a white dress. There are also two motorcycles with riders wearing helmets. The background features tall electrical towers and power lines, as well as some trees and buildings. The sky is clear with a few clouds.\\
  
  \hrulefill
  \medskip
  
  \textbf{\deepseekvthree (Judge):} No The model response describes a general scene of a busy street but fails to mention the core dangerous event: a motorbike riding directly toward the dash-cam vehicle in the wrong direction. Instead, it provides a vague overview of the surroundings without highlighting the unusual or dangerous action that defines the ground truth.
\end{AIboxFull}
\caption{\crashchat fails to detect the motorcycle ahead which is driving against traffic.}
\label{fig:fail_ex}
\end{figure*}

\begin{figure*}[h!]
\centering
\begin{AIboxFull}{\geminiflash fails to describe the motorcycle driving against traffic and instead hallucinates about an action of a car which is driving normally out of the frames.}
  \begin{center}
  \includegraphics[width=0.50\textwidth]{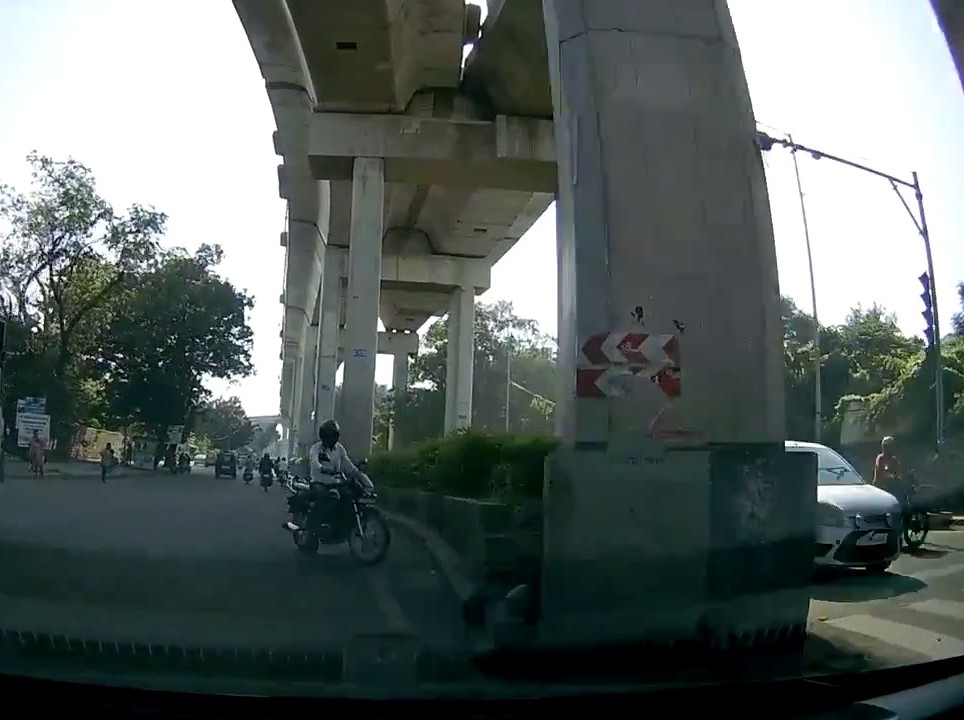}%
  \hfill
  \includegraphics[width=0.50\textwidth]{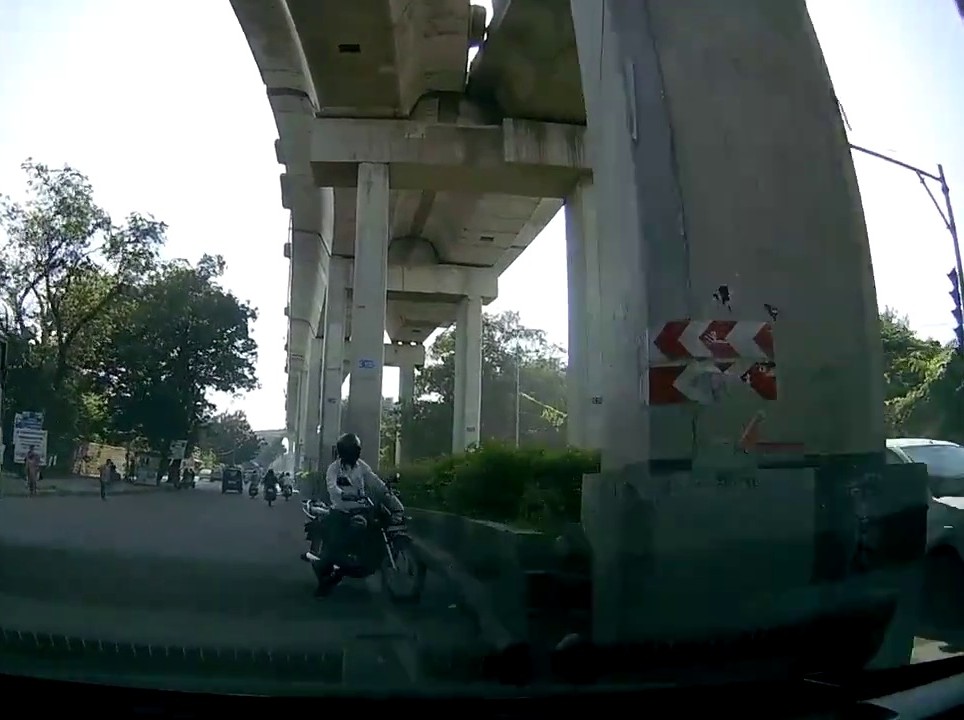}%
  \hfill
  \vspace{-1pt}
  \includegraphics[width=0.50\textwidth]{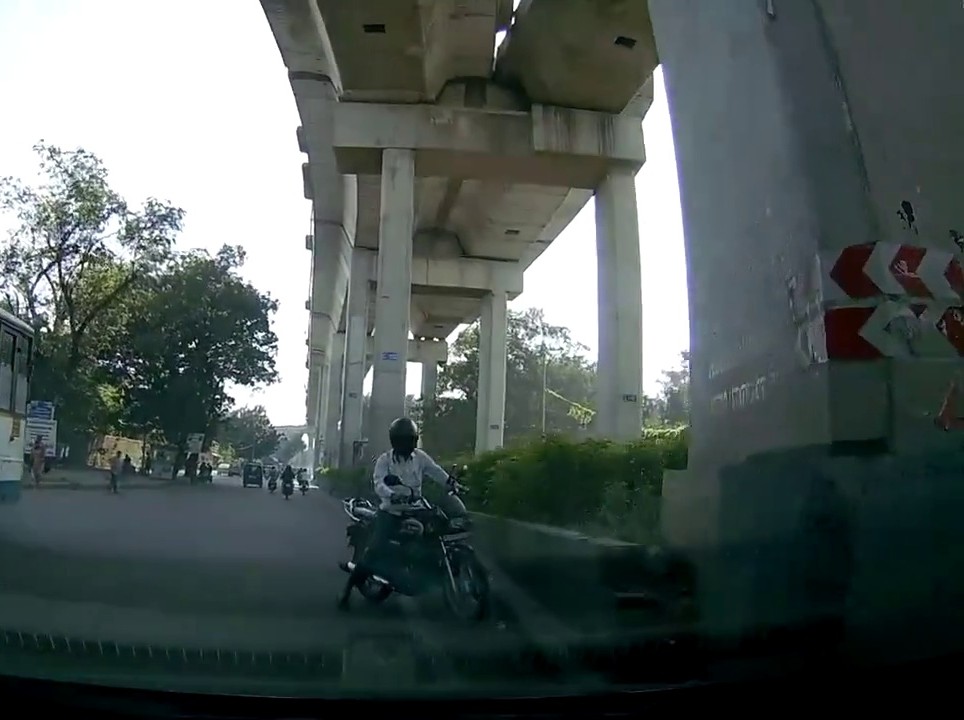}%
  \hfill
  \includegraphics[width=0.50\textwidth]{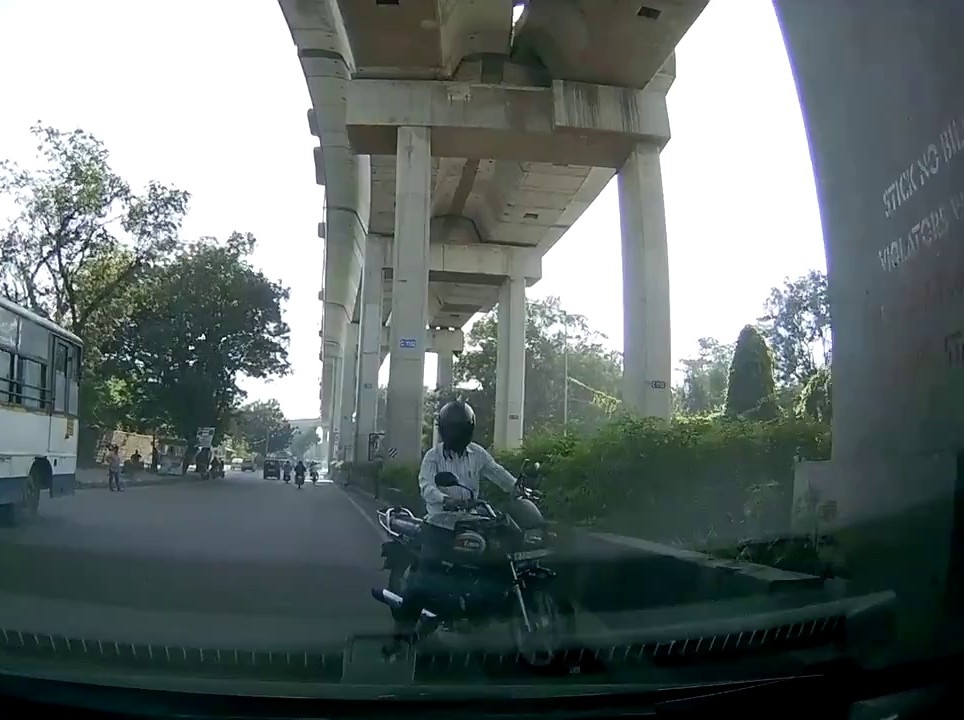}%
  \end{center}
  \medskip

  \textbf{Question Type:} \qone \\
  \textbf{Ground Truth:} A motorbike is riding the wrong way, heading straight against the flow of oncoming traffic.\\
  \textbf{\geminiflash:} The unusual occurrence in this sequence is a white car performing a illegal and dangerous U-turn across multiple lanes.\\
  
  \hrulefill
  \medskip
  
  \textbf{\deepseekvthree (Judge):}No. The model describes a white car performing an illegal U-turn and cutting across lanes. However, the ground truth states a motorbike is riding the wrong way against traffic flow. These are fundamentally different events --- the model focuses on a car's dangerous maneuver, while the ground truth centers on a motorbike's unlawful direction. The core cause, agent, and danger are mismatched.
\end{AIboxFull}
\caption{\geminiflash fails to identify the evident corner case and hallucinates actions of other road agents.}
\label{fig:hallucinate_action_ex2}
\end{figure*}

Despite being evident in the sequence of frames, we show that Video LLMs regardless of proprietary status, size, and whether it was fine-tuned on driving data fails to identify obvious corner cases, instead focusing on other parts of the road, as shown by the response of \crashchat on Figure \ref{fig:fail_ex} and \geminiflash on Figure \ref{fig:hallucinate_action_ex2}.
\subsection{Hallucinating On the Road}
\label{app:hallucinate_ex}
\begin{figure*}[h!]
\centering
\begin{AIboxFull}{\internvlthreefive \texttt{241B} describes a nonexistent cyclist on the road}
  \begin{center}
  \includegraphics[width=0.50\textwidth]{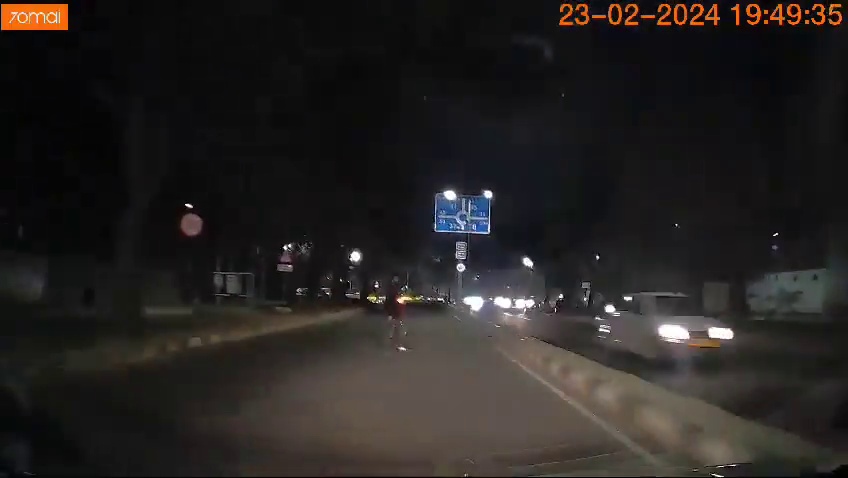}%
  \hfill
  \includegraphics[width=0.50\textwidth]{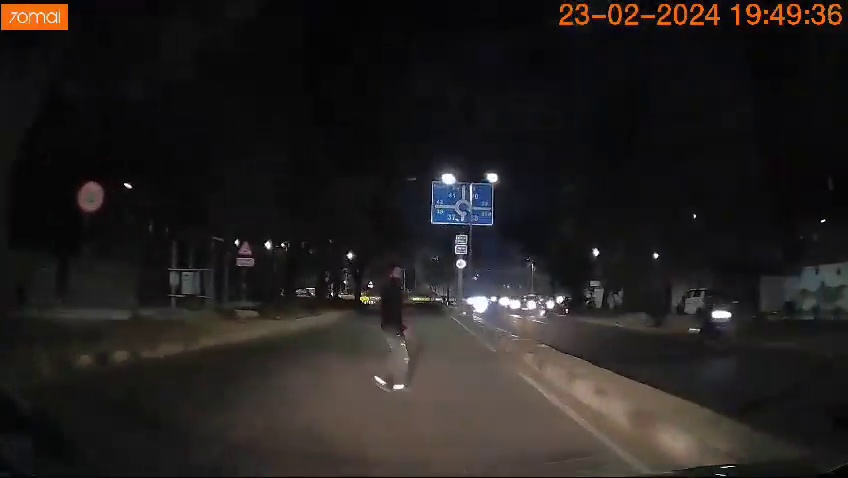}%
  \hfill
  \vspace{-1pt}
  \includegraphics[width=0.50\textwidth]{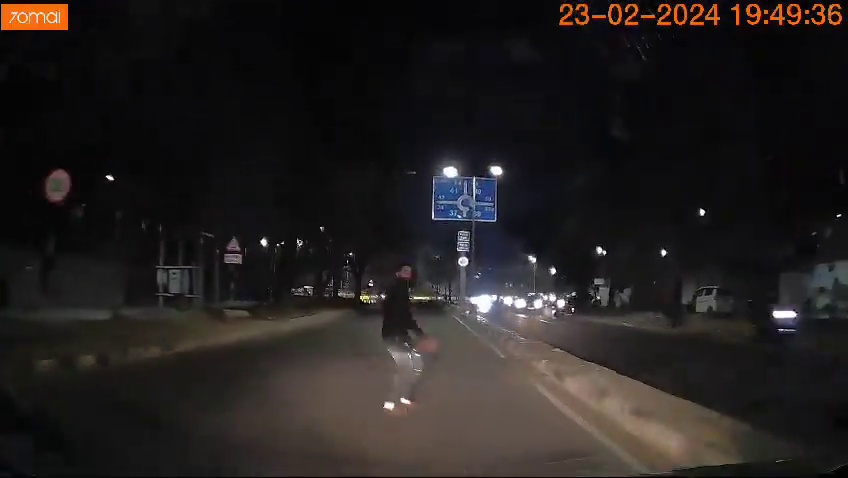}%
  \hfill
  \includegraphics[width=0.50\textwidth]{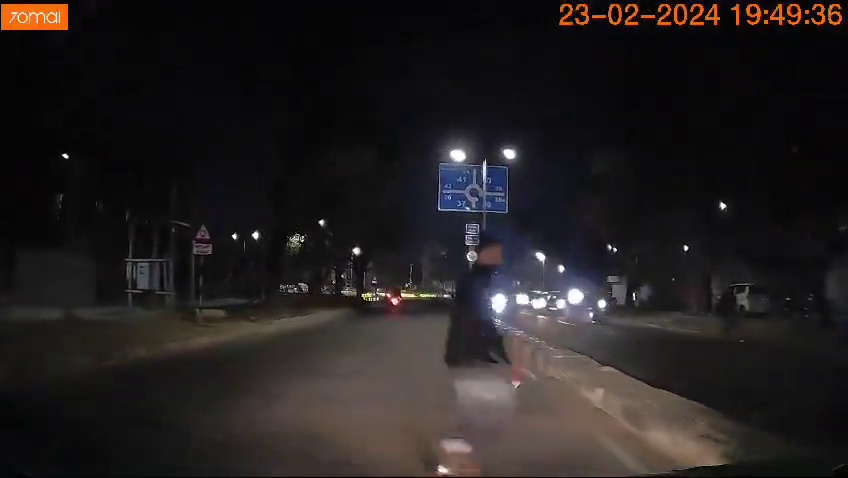}%
  \end{center}
  \medskip

  \textbf{Question Type:} \qtwo \\
  \textbf{Ground Truth:} A pedestrian suddenly steps into the travel lane right in front of the approaching car at night, creating an immediate collision hazard.\\
  \textbf{\internvlthreefive \texttt{241B}:} In this sequence of frames, a person on a bicycle appears suddenly, crossing the road from the right to the left. This could be unexpected for the driver, especially at night when visibility is reduced. The cyclist is not using any lights, making them harder to see. The vehicle continues moving forward without any apparent reaction, which could be risky. It's important for drivers to be cautious and prepared for unexpected obstacles, especially at night.\\
  
  \hrulefill
  \medskip
  
  \textbf{\deepseekvthree (Judge):}No The model describes a cyclist crossing the road unexpectedly, while the ground truth specifies a pedestrian stepping into the lane. The core agents (cyclist vs. pedestrian) and their actions (crossing vs. stepping into the lane) differ significantly, leading to a mismatch in the dangerous event.
\end{AIboxFull}
\caption{\internvlthreefive \texttt{241B} hallucinates about a cyclist on the road when the real danger is a pedestrian hazard.}
\label{fig:hallucinate_ex}
\end{figure*}

Figures \ref{fig:hallucinate_ex} shows a case where Video LLMs hallucinate road agents or objects on the road, which are simply not there. For possible deployment as part of a vision-language-action framework enabler, Video LLMs need to overcome this glaring issue of hallucinating on the road.

\subsection{Collisions as Near-Misses}
\label{app:coll_ex}
\begin{figure*}[h!]
\centering
\begin{AIboxFull}{\glmfouronevthinking cannot distinguish between an evident collision and a near-miss}
  \begin{center}
  \includegraphics[width=0.50\textwidth]{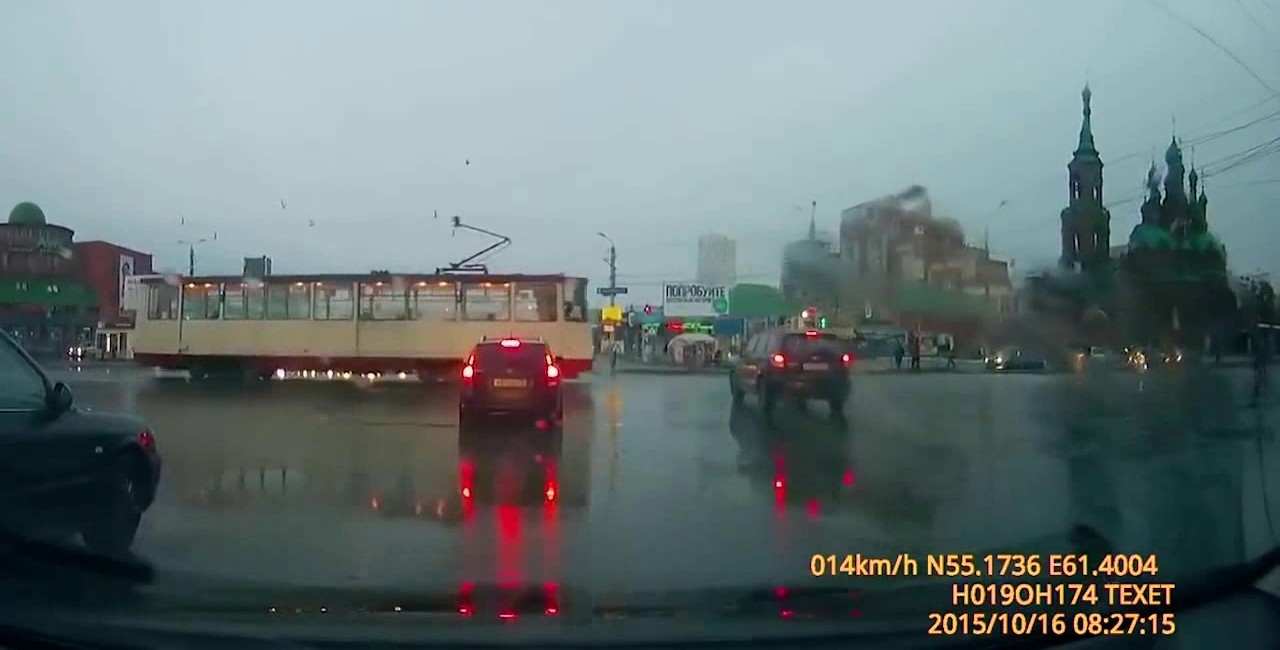}%
  \hfill
  \includegraphics[width=0.50\textwidth]{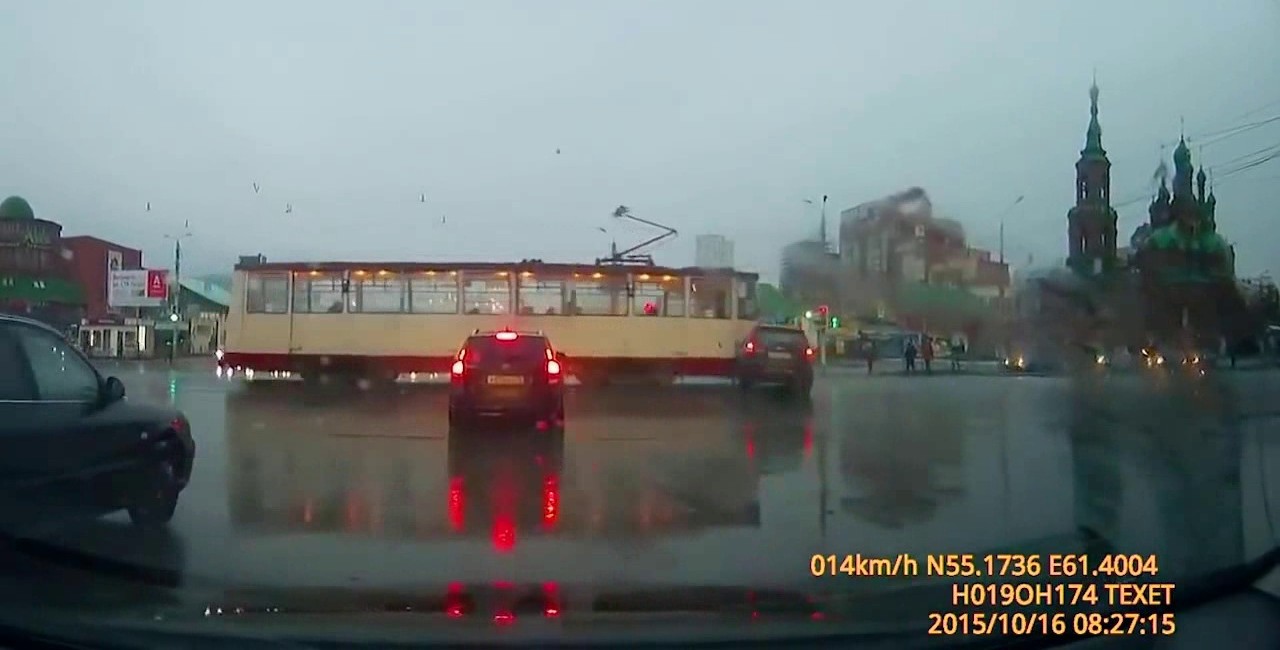}%
  \hfill
  \vspace{-1pt}
  \includegraphics[width=0.50\textwidth]{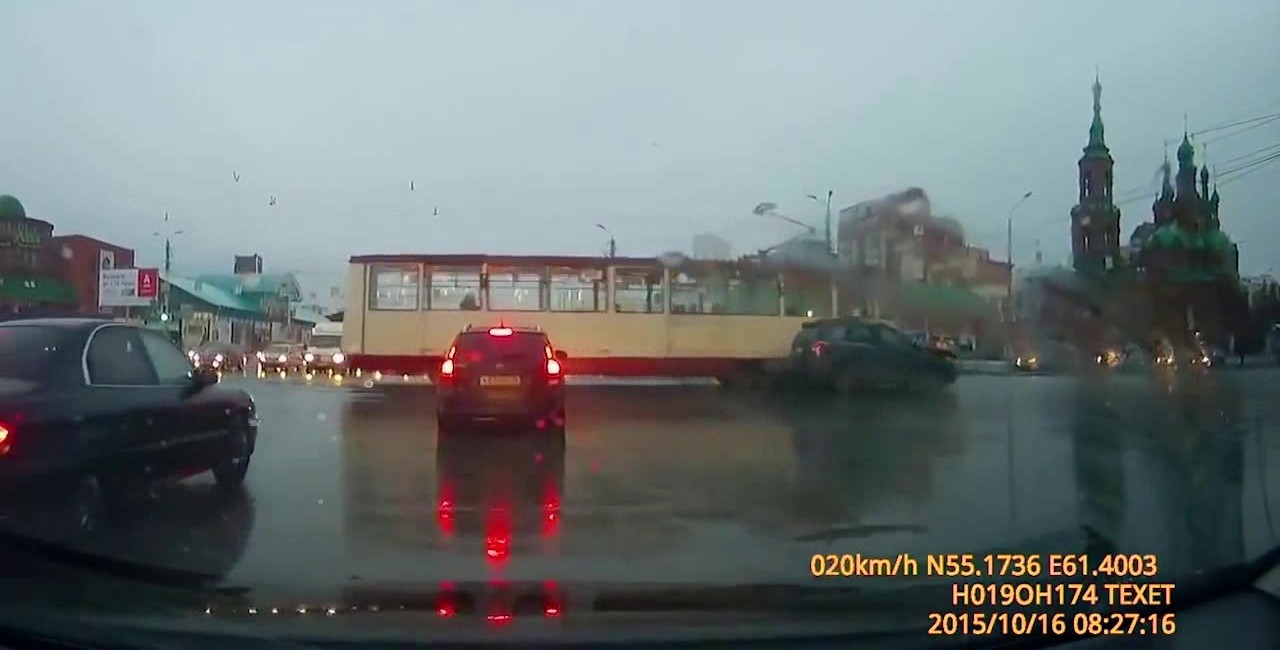}%
  \hfill
  \includegraphics[width=0.50\textwidth]{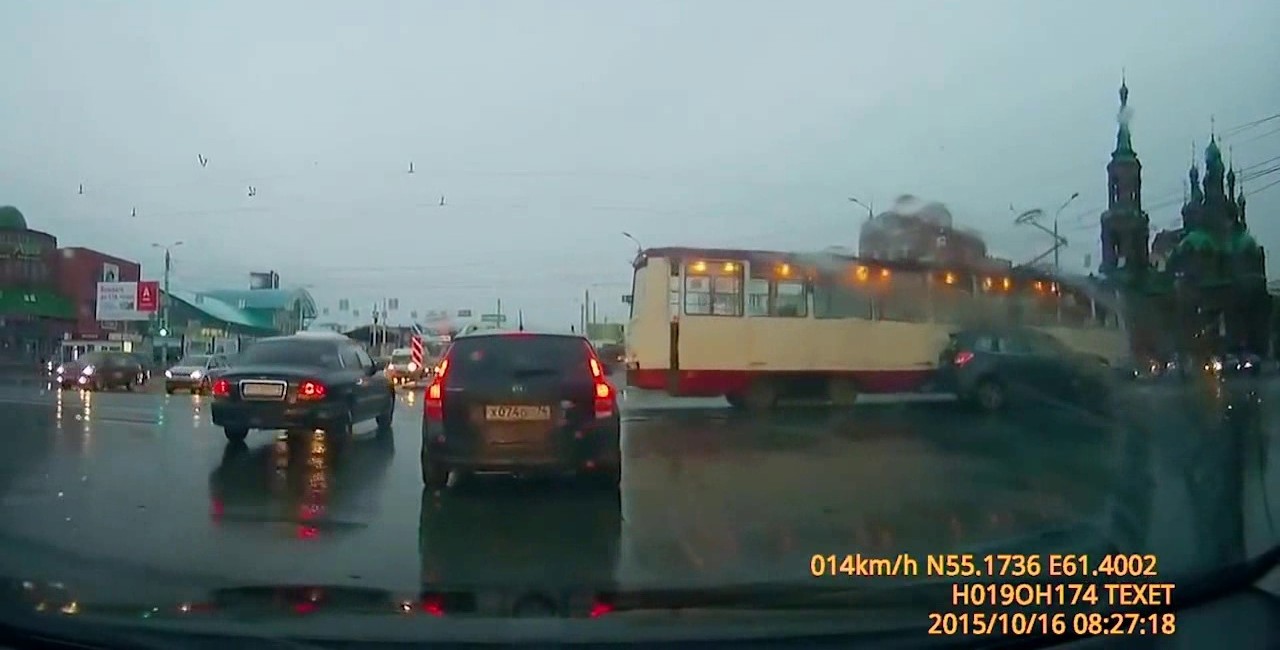}%
  \end{center}
  \medskip

  \textbf{Question Type:} \qone \\
  \textbf{Ground Truth:} A car drives into the side of a tram as it crosses the intersection.\\
  \textbf{\glmfouronevthinking} The unusual element is likely a **near - collision** (common in such busy, wet conditions), as the scene involves multiple vehicles (cars, tram) in close proximity with potential unsafe maneuvers. The rainy weather and dense traffic compound the risk, making the sequence\u2019s chaotic interaction between vehicles (especially the tram and cars) the unusual factor. In summary, the unusual aspect is a near - collision in the rainy, busy intersection involving multiple vehicles (including the tram).\\
  
  \hrulefill
  \medskip
  
  \textbf{\deepseekvthree (Judge): }No The model response vaguely describes a \"likely near-collision\" involving multiple vehicles in rainy conditions but does not specify the core event: a car driving into the side of a tram. The ground truth clearly identifies the exact dangerous interaction (car hitting tram), while the model only infers a general risk scenario without pinpointing the specific cause or agents involved. The temporal and causal precision is missing.
\end{AIboxFull}
\caption{\glmfouronevthinking fails to confidently claim that a collision is occurring, demonstrating lack of visual prowess.}
\label{fig:coll_ex}
\end{figure*}

Genuine collisions on the road are important corner cases because they might affect subsequent actions that will be taken by the ADS based on the Video LLM's observation ; treating them as near-misses due to failing to localize the exact moment where a collision occurs (possibly due to the large number of frames passed) presents a severe issue that should be addressed. Figure \ref{fig:coll_ex} shows this occurrence.

\subsection{Making Up Actions on the Road}
\label{app:hallucinate_action_ex}
\begin{figure*}[h!]
\centering
\begin{AIboxFull}{\llavanextvideo misses a motorcycle collision and talks about a car running a red light instead}
  \begin{center}
  \includegraphics[width=0.50\textwidth]{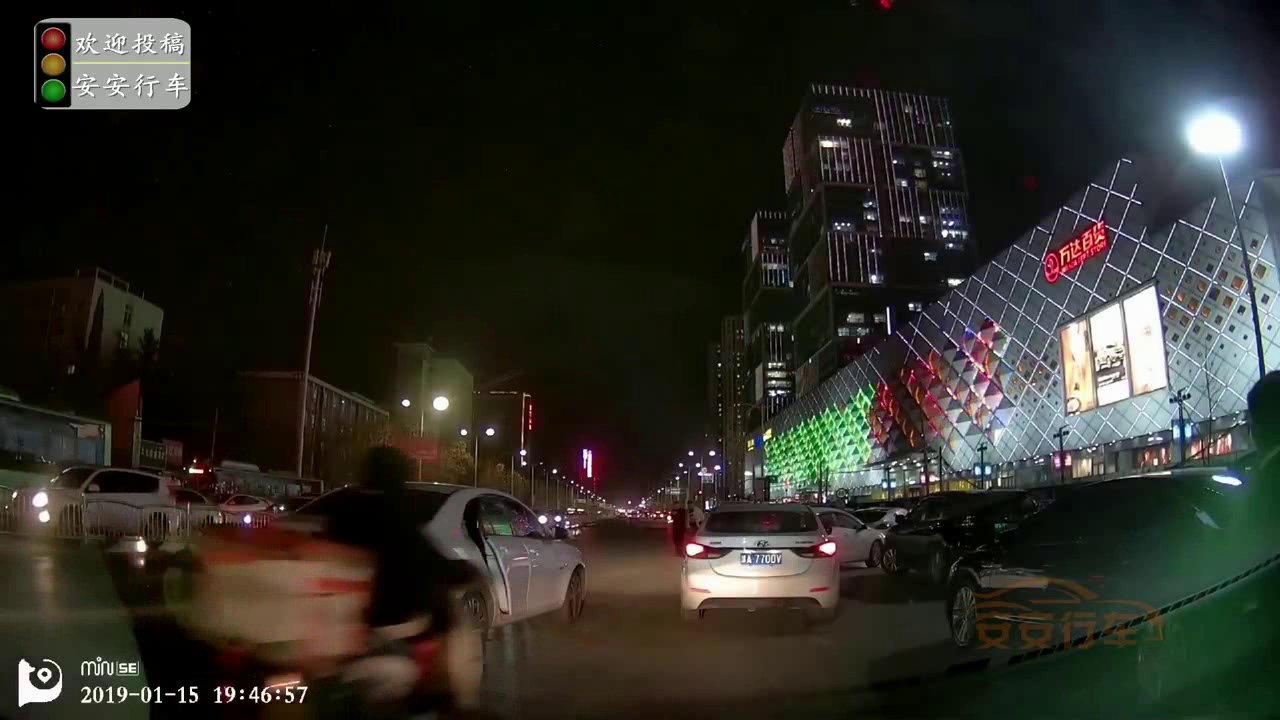}%
  \hfill
  \includegraphics[width=0.50\textwidth]{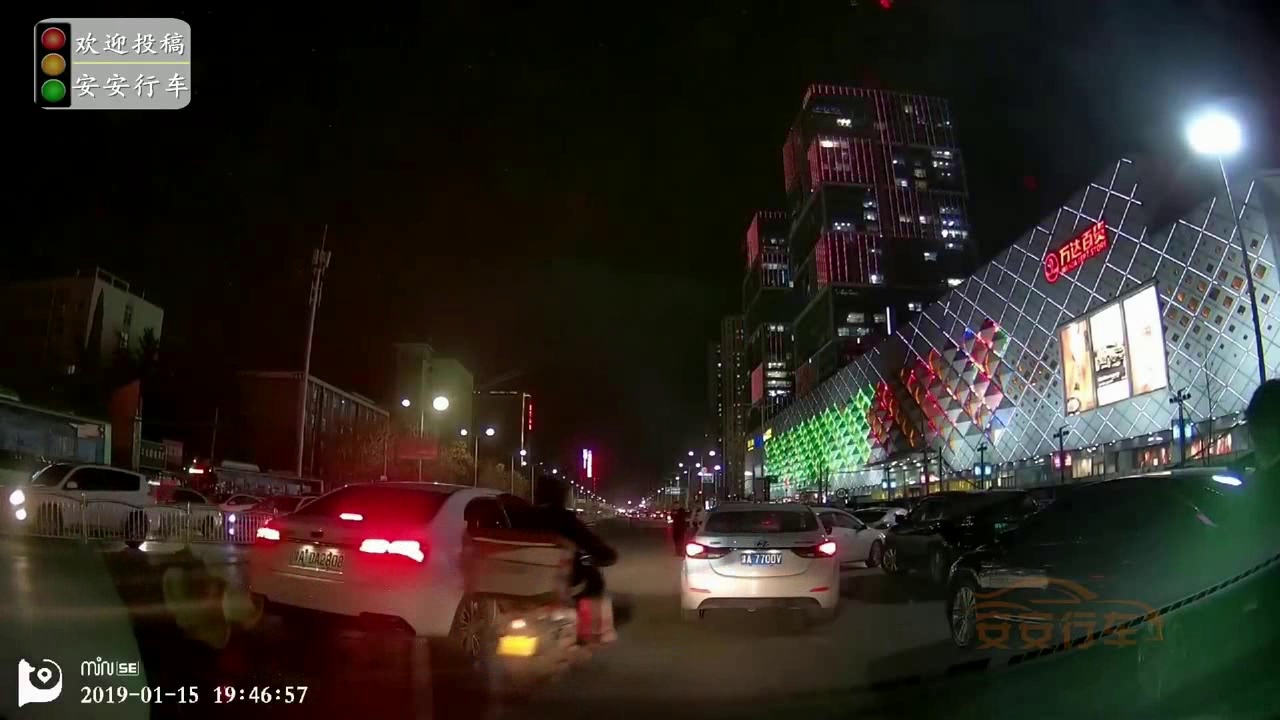}%
  \hfill
  \vspace{-1pt}
  \includegraphics[width=0.50\textwidth]{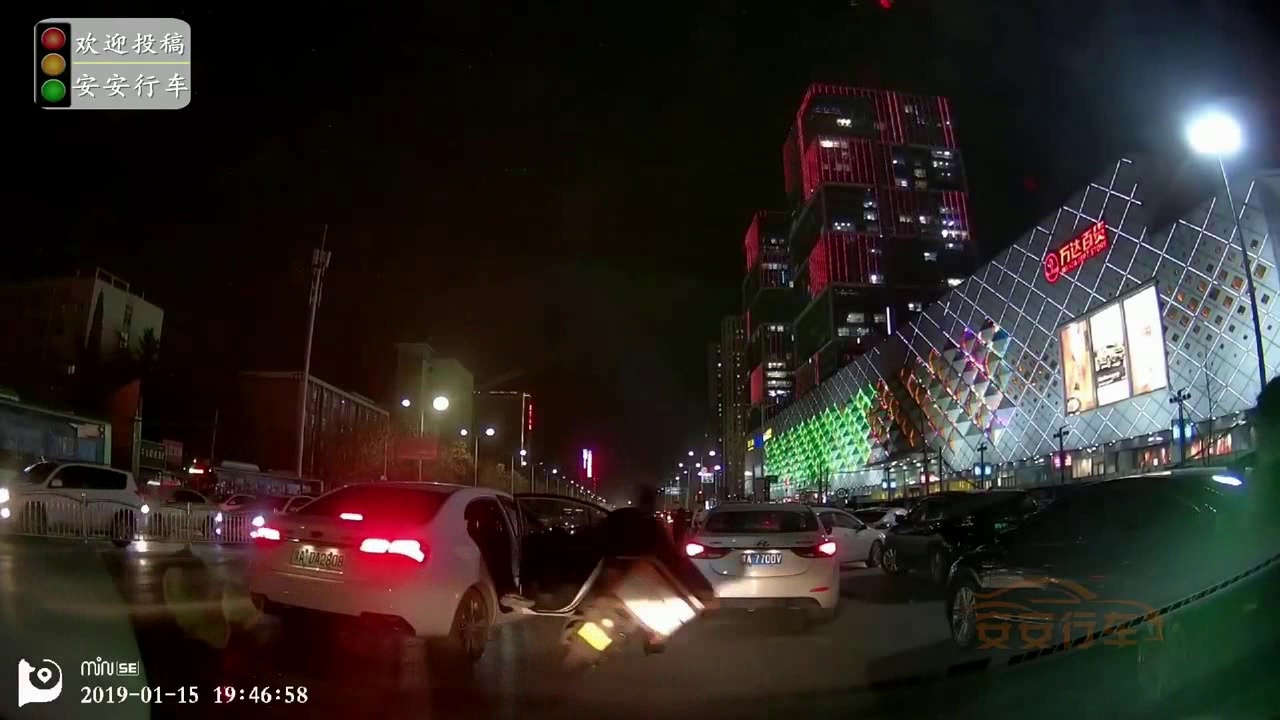}%
  \hfill
  \includegraphics[width=0.50\textwidth]{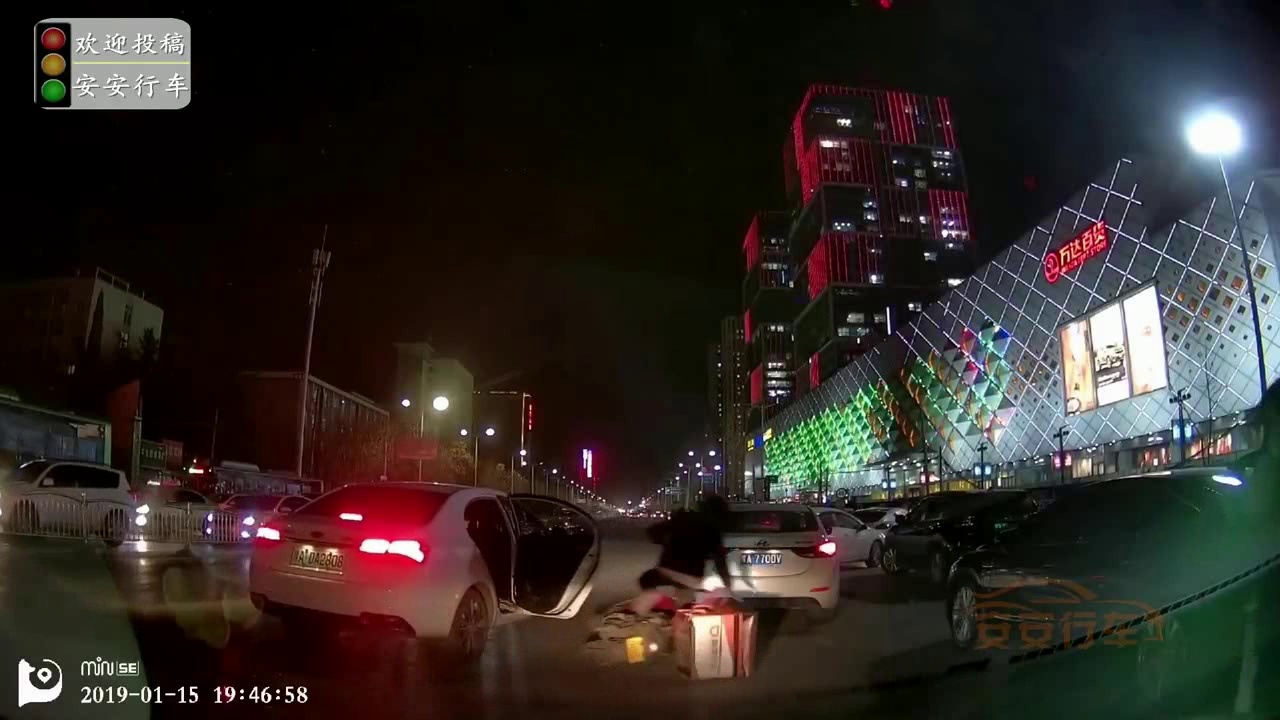}%
  \end{center}
  \medskip

  \textbf{Question Type:} \qfive \\
  \textbf{Ground Truth:} A motorbike collides with the suddenly opened door of a white car, scraping it and throwing the rider to the ground.\\
  \textbf{\llavanextvideo:} The video shows a corner case where a car is seen running a red light and colliding with another vehicle. This is an unexpected and unusual driving situation that deviates from normal traffic behavior, as running a red light is generally against traffic rules and can lead to accidents. The car that runs the red light is the one that causes the collision, and the other car is the one that gets hit. This is a rare and potentially hazardous event that deviates from the normal flow of traffic and can lead to accidents.\\
  
  \hrulefill
  \medskip
  
  \textbf{\deepseekvthree (Judge):}No The model describes a car running a red light and colliding with another vehicle, while the ground truth involves a motorbike hitting a suddenly opened car door, which is a completely different scenario involving different agents, causes, and effects.
\end{AIboxFull}
\caption{\llavanextvideo could not identify the motorcycle hitting the car and hallucinates a car running a red light.}
\label{fig:hallucinate_action_ex}
\end{figure*}

Another concerning point for Video LLMs as future co-navigators and planners is if they invent or misinterpret the actions of other road agents (e.g., a swerving car when in reality the car is driving straight). Figure \ref{fig:hallucinate_action_ex} shows a case where the Video LLMs' hallucination of actions and behavior can lead to different crucial decisions made in that scenario.

\section{License and Dataset Usage}
\label{appsec:license}

DashBench is derived from two publicly available datasets and is intended strictly for 
non-commercial research use.

\textbf{MM-AU}~\cite{DBLP:conf/cvpr/FangLZXYLXC24} is released for non-commercial use only, covering educational 
and research purposes. MM-AU itself aggregates footage from publicly available ego-view accident datasets and video platforms (YouTube, 
Bilibili, Tencent), with all original rights remaining with their respective owners.

\textbf{RoadSocial}~\cite{DBLP:conf/cvpr/ParikhRTGS25} is released under the \textbf{Creative Commons 
Attribution--NonCommercial--ShareAlike 4.0 International (CC BY-NC-SA 4.0)} license. 
The dataset authors do not claim ownership of the linked video content; all rights 
and copyright remain with the respective content creators and platform owners. Users 
are responsible for ensuring compliance with the terms of the hosting platforms.

\end{document}